\documentclass[10pt,twocolumn,letterpaper]{article}

\usepackage[pagenumbers]{cvpr}

\usepackage[dvipsnames]{xcolor}

\usepackage{times}
\usepackage{epsfig}
\usepackage{graphicx}
\usepackage{amsmath}
\usepackage{amssymb}
\usepackage[dvipsnames]{xcolor}
\usepackage{bm}
\usepackage{caption}
\usepackage{fontawesome}
\usepackage{subcaption}
\usepackage{multirow}
\usepackage{soul}
\usepackage{colortbl}

\usepackage{array}
\newcolumntype{H}{>{\setbox0=\hbox\bgroup}c<{\egroup}@{}}
\newcolumntype{P}[1]{>{\centering\arraybackslash}p{#1}}

\definecolor{cvprblue}{rgb}{0.21,0.49,0.74}
\usepackage[pagebackref,breaklinks,colorlinks,citecolor=cvprblue]{hyperref}

\newcommand{\OursBody}[0]{CroCo-Body\xspace}
\newcommand{\OursHand}[0]{CroCo-Hand\xspace}
\newcommand{\hlcell}[0]{\cellcolor{gray!22}} 

\title{Cross-view and Cross-pose Completion for 3D Human Understanding}

\newcolumntype{x}[1]{>{\centering\arraybackslash\hspace{0pt}}p{#1}}
\author{
\begin{tabular}{x{3.7cm}@{}x{3.7cm}@{}x{3.7cm}@{}x{3.7cm}}
\\[-0.65cm]
Matthieu Armando & Salma Galaaoui & Fabien Baradel & Thomas Lucas \\
Vincent Leroy & Romain Br\'egier & Philippe Weinzaepfel & Gr\'egory Rogez \\[0.12cm]
\multicolumn{4}{c}{NAVER LABS Europe} \\[-0.12cm]
\multicolumn{4}{c}{\small \url{https://europe.naverlabs.com/ComputerVision/CroCoMan}} \\[-0.2cm]
\end{tabular}
}

\begin{document}

\twocolumn[{
\renewcommand\twocolumn[1][]{#1}
\maketitle
\begin{center}
    \centering
    \vspace{-0.6cm}
    \includegraphics[width=\linewidth]{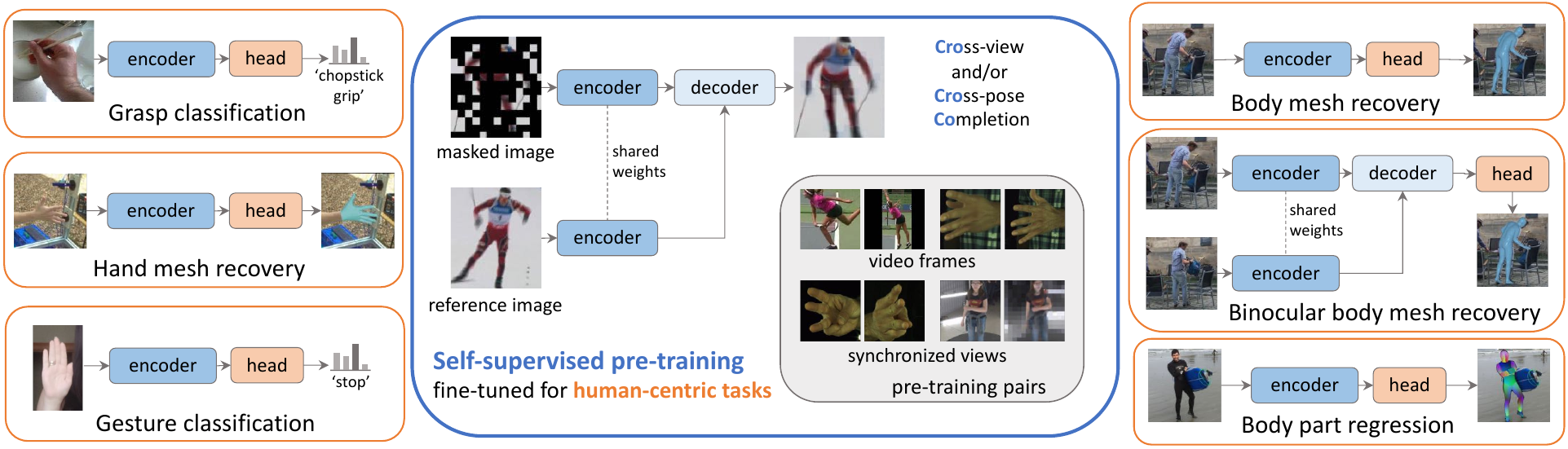} \\[-0.3cm]
    \captionof{figure}{\small{\textbf{Human-centric pre-training.} We pre-train a model for cross-view and cross-pose completion 
    on body and hands image pairs (middle). This model serves as initialization for fine-tuning on several downstream tasks 
    for both hands (left) and bodies (right).
    Our model, based on a generic transformer architecture, achieves competitive performance on these tasks without bells and whistles. 
    }}
    \label{fig:teaser}
    \vspace{0.1cm}
\end{center}
}]

\begin{abstract}
Human perception and understanding is a major domain of computer vision which, like many other vision subdomains recently, stands to gain from the use of large models pre-trained on large datasets. We hypothesize that the most common pre-training strategy of relying on general purpose, object-centric image datasets such as ImageNet, is limited by an important domain shift. On the other hand, collecting domain-specific ground truth such as 2D or 3D labels does not scale well. Therefore, we propose a pre-training approach based on self-supervised learning that works on human-centric data using only images. Our method uses pairs of images of humans: the first is partially masked and the model is trained to reconstruct the masked parts given the visible ones and a second image. It relies on both stereoscopic (cross-view) pairs, and temporal (cross-pose) pairs taken from videos, in order to learn priors about 3D as well as human motion. We pre-train a model for body-centric tasks and one for hand-centric tasks. With a generic transformer architecture, these models outperform existing self-supervised pre-training methods on a wide set of human-centric downstream tasks, and obtain state-of-the-art performance for instance when fine-tuning for model-based and model-free human mesh recovery.
\end{abstract}

\section{Introduction}

The main catalyst of performance improvement in computer vision tasks in the last decade has arguably been the training of large models on large datasets \cite{lin2014microsoft,deng2009imagenet,carreira2019short,yu2015lsun,Rothe-ICCVW-2015,krishna2017visual,liu2018large}.
For human-centric vision tasks, the standard approach is to pre-train 
models on ImageNet classification tasks and then fine-tune them on downstream tasks with specific datasets \cite{h36m_pami,guler2018densepose,mono-3dhp2017,3dpw}.
This has at least three drawbacks: a) the size of the pre-training dataset is limited by label acquisition, b) there can be a large domain shift between ImageNet and downstream images, c) object-centric classification is different from human understanding, which may hinder the relevance of pre-trained features.
Collecting large annotated datasets for human centric vision tasks is hard: target signal is costly and hard to acquire in the wild, \eg relying on motion capture systems to obtain 3D pose.

\begin{figure*}
    \centering
    \newcommand{\pairwidth}{0.32\linewidth}
    \newcommand{\video}{\faVideoCamera}
    \newcommand{\multiview}{\faCamera}
    \newcommand{\synth}{\faLaptop}
    \newcommand{\film}{\faFilm}
    \begin{tabular}{c@{ }c@{ }c}
    \includegraphics[width=\pairwidth]{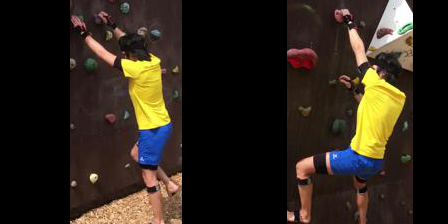}&
    \includegraphics[width=\pairwidth]{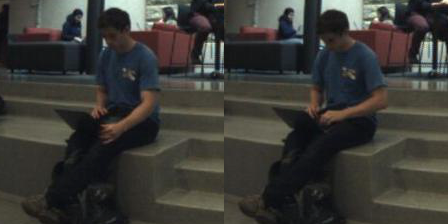}&
    \includegraphics[width=\pairwidth]{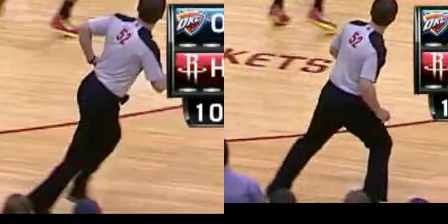} \\[-0.1cm]
3DPW~\cite{3dpw}~\video & JRDB~\cite{jrdb}~\video & PoseTrack~\cite{posetrack}~\video \\[0.1cm]
    \includegraphics[width=\pairwidth]{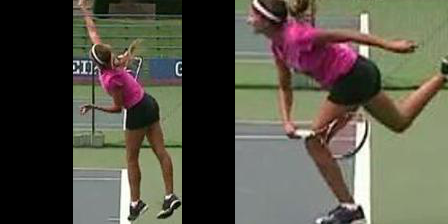}&
    \includegraphics[width=\pairwidth]{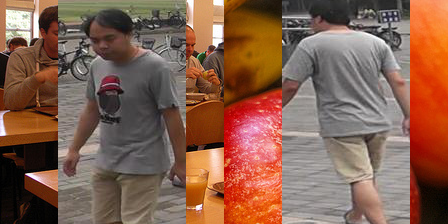}&
    \includegraphics[width=\pairwidth]{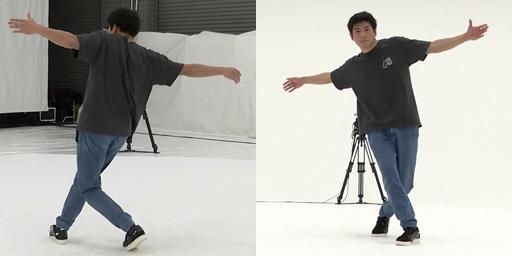} \\[-0.1cm]
    PennAction~\cite{pennaction}~\video & MARS~\cite{mars}~\video & AIST~\cite{aist}~\multiview~\video \\[0.1cm]
    \includegraphics[width=\pairwidth]{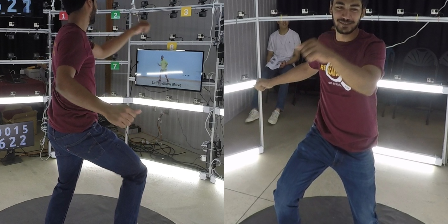}&
    \includegraphics[width=\pairwidth]{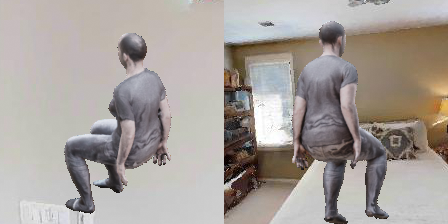} &
    \includegraphics[width=\pairwidth]{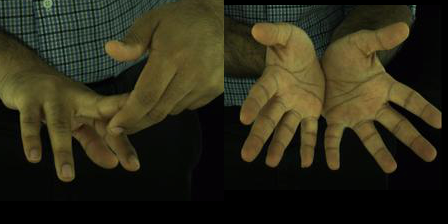} \\[-0.1cm]
    HUMBI~\cite{humbi}~\multiview & synthetic~\synth~\multiview & InterHand2.6M~\cite{interhand}~\multiview~\video \\[0.1cm]
    \end{tabular}    \\[-0.5cm]
    \caption{\textbf{Examples of pre-training pairs} taken from the different pre-training datasets. \multiview~denote multi-view datasets, \video~video datasets and \synth~synthetic data.  \\
    }
    \label{fig:pairs}
    \vspace*{-0.8cm}
\end{figure*}

To leverage large amounts of data and scale to large models, self-supervised pre-training methods such as contrastive learning \cite{wu2018npid,caron2018deep,simclr,byol,dino} and masked signal modeling \cite{BERT,mae} -- have been developed. In these paradigms, a  pretext task is constructed from the data itself so that no manual labeling is required. The epitome of this philosophy is that of foundation models, such as GPT \cite{brown2020language}, trained on a large corpus of data with no manual annotation and fine-tuned to perform efficiently various downstream tasks. 
In computer vision, most self-supervised learning methods have been developed on ImageNet and evaluated chiefly on image classification tasks. Pioneering work in~\cite{vitpose} provided a study of masked image modeling (MIM) pre-training on data dedicated to human-centric tasks, but found limited success.
In contrast, we design a pre-training method to capture prior knowledge about the human body, which we use with human-centric data -- see Figure~\ref{fig:teaser} for an overview.
It is inspired by masked image modeling \cite{BERT,mae}, where parts of an image are hidden and the network is trained to reconstruct them using the visible ones as input. 

Unlike MAE~\cite{mae} which operates on individual images, in our case  pairs of images of human bodies are leveraged. 
These pairs are constructed in two ways: a) by taking two views of the same pose -- plenty of those are available but the variability in pose and background is limited \cite{aist,humbi,interhand} -- and b) by taking two poses in a motion sequence at different time steps -- for instance from videos showing a person in movement, which are also plentiful \cite{3dpw,jrdb,pennaction,posetrack,mars}; see examples in Figure~\ref{fig:pairs}.
Once a pair is constructed, parts of the first pose are masked, and the network is trained to reconstruct them from the unmasked regions as well as from the second image. 
This approach is close in spirit to cross-view completion (CroCo), proposed in~\cite{croco}, which performs masked image modeling with pairs of images. 
However, CroCo works with a different type of data (typically buildings or interiors~\cite{habitat}) with relatively small view-point changes. 
In their case, the objects are rigid, and therefore CroCo only requires static pairs of the first type.  
To capture the deformable nature of the human body, we generalize their approach to the second type of pairs, --- two different poses of a dynamic motion -- and we refer to the corresponding task as cross-pose completion. We also include static pairs with extreme viewpoint changes, such as front and back views, to acquire a broad 3D understanding of the human body beyond stereo reconstruction.
We propose a procedure to build suitable pairs of both types. These are taken from two human pose and motion modalities: views of full human bodies, and closeups of hands. This covers a wider input domain and captures information about human motion at two levels. Indeed, one of the specificity of general human-centric vision is the wide range of expected accuracy depending on the task. For instance, images of humans from afar may be sufficient for body pose estimation, but the millimetric accuracy expected for hand pose estimation requires close-ups of these parts.

Empirically, we pre-train a transformer architecture using the cross-pose and cross-view completion pretext tasks, on the two data modalities, \ie, human bodies and hands.
We fine-tune our model on the human-centric downstream tasks of pose and mesh recovery for bodies and hands~\cite{kolotouros2019learning,joo2021exemplar}, and dense pose estimation~\cite{guler2018densepose, neverova2020continuous}. 
Using a generic transformer-based architecture, we demonstrate that our pre-training method transfers better than supervised ImageNet pre-training or existing self-supervised pre-training methods when fine-tuning on downstream tasks.
This allows our models to achieve competitive performance on these downstream tasks without requiring task-specific designs.
We release  two pre-trained models for human-centric vision:~\OursBody and~\OursHand, specialized either for body or hand related tasks respectively. We demonstrate that these can easily be transferred to a variety of coarse- and fine-grained downstream tasks. 
For instance, our pre-training allows reaching state-of-the-art performance for model-based and model-free human mesh recovery.


\section{Related work}

We first discuss self-supervised learning in computer vision and then their application to human-centric tasks.

\noindent \textbf{Self-supervised learning.} 
Popular methods for self-supervised pre-training can be coarsely separated into two categories.
The first one is based on instance discrimination~\cite{wu2018npid,simclr,byol,swav,dino} where  various data augmentations are applied to an image to obtain different variants of it, and features extracted from the model are trained to be similar for these variants while being different from features extracted from other images.
These methods produce models that transfer remarkably well to image-level tasks such as image classification and have achieved state-of-the-art performance on several vision benchmarks. However, recent studies suggest that the \textit{object-centric} and the \textit{balanced} nature of ImageNet~\cite{purushwalkam2020demystifying,assran2022hidden,imagenet} play a big role in this success, and indeed transfer performance degrades when tasks of a very different nature are considered~\cite{van2018inaturalist}. 

The second paradigm relies on masked image modeling (MIM). Inspired by BERT~\cite{BERT} in natural language processing,
MIM aims at reconstructing masked information from an input image either in pixel space~\cite{mae,SimMIM,sit,iGPT,splitmask,multimae} or in feature space~\cite{data2vec,maskfeat,msn}. 
These methods have obtained some success on denser tasks such as object detection~\cite{mae} and human pose estimation~\cite{vitpose}, which is of particular interest to our work.
In~\cite{croco}, a MIM method that works with pairs of scene images has been proposed based on the cross-view completion task (CroCo). 
A second view of a scene is added to the MIM framework: this view can be used to improve reconstruction accuracy of the masked parts, provided that the model can exploit scene geometry. 
This approach allows the design of pretext tasks through a careful choice of image pairs. 
While CroCo is designed for static problems, we design a pre-training procedure intended to capture the articulated nature of the human body, by building pairs taking different poses from a dynamic motion, referred to as cross-pose completion, which we use alongside cross-view completion.
MIM has also been applied on videos~\cite{videomae,spatioMae}: information is also shared between frames, though in this case the goal is to learn from context rather than geometry.

\noindent \textbf{Human-centric pre-training.}
A large percentage of images/videos captured in our daily life contains humans.
However collecting annotations related to human pose in an image is cumbersome, especially when 3D annotations are required~\cite{hampali2020honnotate,3dpw}, and in practice supervised data is scarce. This 
motivates the development of self-supervised methods tailored to human-centric data. 
In recent works~\cite{zhao2021learning,liu2022temporal,vitpose}, existing self-supervised learning strategies were adapted to human-centric data. A popular approach  is to leverage the fact that humans have a well-defined structure, for instance by learning human pose in an unsupervised manner~\cite{rhodin2018unsupervised,schmidtke2021unsupervised,jakab2018unsupervised,jakab2020self}. 
Jakab \etal~\cite{jakab2018unsupervised} extract 2D landmark of faces and bodies in an unsupervised way by learning to disentangle pose from appearance through conditional image generation from image pairs. However, the learned 2D landmarks do not have structure and do not correspond to any predefined human skeleton. 
In their following work~\cite{jakab2020self}, this issue was solved 
by leveraging Motion Capture (MoCap) data to ensure that the 2D landmarks are matching a certain pose distribution. However, this is quite limiting, especially in the presence of rare or unseen poses.
Following a different pre-training strategy, HCMoCo~\cite{hong2022hcmoco} learn a representation on multimodal data including RGB, depth or 2D keypoints, using contrastive learning. It obtains convincing results on multiple downstream tasks such as part segmentation or 3D pose estimation.
However, acquiring such multimodal data outside of studio environments is difficult and prevents the use of this method in the wild. 
UniHCP~\cite{ci2023unihcp} and PATH~\cite{tang2023humanbench} followed this direction by both proposing to learn a model using multitask training. They required annotations on several tasks however, and mainly focused on 2D downstream tasks.
Most recently, LiftedCL~\cite{liftedcl} leverages contrastive learning to obtain a representation invariant to data augmentations.
They encourage the pose representation to be realistic via the use of an adversarial strategy on 3D skeleton data.
More recently,
SOLIDER~\cite{chen2023beyond} focused on disentangling semantics and appearances in an unsupervised manner.
They generate pseudo semantic labels using DINO~\cite{dino}.
While reaching good performances on 2D tasks such as human parsing or attribute recognition, they do not present results on 3D tasks.

 Contrary to these, we do not 
impose any constraint on the learned human representation.
We instead cast our pre-training as a conditional image generation where image pairs are selected from different 
viewpoints or time steps.

\section{Method}

We first present our pre-training objective (Section~\ref{subsec:MIM}), then our data collection process to collect cross-view (Section~\ref{subsec:crossview}) and cross-pose (Section~\ref{subsec:crosspose}) data, and finally the fine-tuning on downstream tasks (Section~\ref{subsec:finetune}).

\subsection{Multi-view masked image modeling} \label{subsec:MIM}
Masked image modeling as proposed in \cite{mae} proceeds by dividing an image $\bm{x}$ into $N$ non-overlapping patches $\bm{p}= \{\bm{p}^1,\hdots,\bm{p}^{N}\}$.  A subset of $n=\lfloor rN \rfloor$ tokens (\ie, encoded patches) is randomly masked, with $r\in[0,1]$ an hyper-parameter controlling the masking ratio, and the remaining set of visible patches is denoted 
$\bm{\tilde{p}} = \{\bm{p}^i | m_i=0\}$, with $m_i=0$ indicating that the patch $\bm{p_1}^i$ is not masked ($m_i=1$ otherwise).
A model is trained to predict masked tokens given the visible ones.
While there is an inherent ambiguity on the masked content, empirically \cite{mae} has shown that such an approach of minimizing a deterministic $L_2$ loss between prediction and target is a succesful pre-training objective.    

The framework proposed in CroCo~\cite{croco} extends this concept to pairs of images. In our case, we build two types of image pairs: a) cross-pose; where the pair comes from two different time-steps but  the same camera or -- b) cross-view; where the pair depicts the same pose of a person seen from two different viewpoints, see Figure~\ref{fig:pairs} for examples of pairs. Let $x^v_t$ be an image viewed from viewpoint $v$ and taken at timestep $t$. Images in a pair 
are divided into sets of patches, denoted $(\bm{p_{t}^v}, \bm{p_{t'}^w})$, patches from the target are randomly masked, let $\bm{\tilde{p}_{t}^v}$ denote the set of remaining patches, both views are encoded separately with an encoder $E_\theta$ sharing the weights between the two images, and decoded together with a decoder 
 $D_\phi$:
\begin{equation}
\bm{\hat{p_t^v}}_{\theta, \phi} = D_{\phi}\left(E_\theta(\bm{\tilde{p}_{t}^v});E_\theta(\bm{p_{t'}^w})\right).
\end{equation}

While CroCo considered \emph{static} pairs of images, denoted  $(\bm{x}^{v}, \bm{x}^{w})$ in this work, the human body is a deformable object; therefore we also consider \emph{dynamic} pairs, representing different poses taken at different steps of a human motion, denoted $(\bm{x}_{t}, \bm{x}_{t'})$. We denote $\mathcal{D}_\text{pose}$ the set of cross-pose (dynamic) pairs and $\mathcal{D}_\text{view}$ the set of cross-view (static) image pairs. The model is trained by optimizing:
\begin{equation}
\resizebox{\linewidth}{!}{$
\underset{\theta, \phi}{\texttt{min}} \:\: \underbrace{\sum_{\small (\bm{p}_t, \bm{p}_{t'}) \in \mathcal{D}_\text{pose}} \vert\vert \hat{\bm{p}_t}_{\theta, \phi} - \bm{p}_t \vert\vert^2}_{\text{(a) cross-pose completion}} + \underbrace{\sum_{\small (\bm{p^v}, \bm{p^w}) \in \mathcal{D}_\text{view}} \vert\vert \hat{\bm{p^v}}_{\theta, \phi} - \bm{p^v} \vert\vert^2.}_{\text{(b) cross-view completion}}$
}
\end{equation} 

In the case where only (b) is used, the objective boils down to the same as in  \cite{croco}, with different data. 

\noindent \textbf{Human-centric masking strategy.}
Following standard MIM practice, we mask a subset of whole tokens (\ie, squared patches in the image).
While standard MIM methods mask each token with equal probability, we leverage
the prior knowledge that pixels belonging to the humans are the relevant ones. 
Therefore, when segmentations can be easily obtained, we explore non-uniform masking strategies. 
Specifically, the set of $N$ image patches is separated into $N_H$ \textit{human} and $N_B$ \textit{background} patches ($N = N_H + N_B$). The masking ratio is applied to $N_H$ only, \ie we keep visible $n=N_H - \lfloor rN_H \rfloor$ patches.
Since $N_H$ varies between images, so does $n$: for each training batch, we add to the visible set of patches randomly chosen background patches, up to a fixed sequence length set to the maximum of $N_H$ over the batch.
This has two consequences: a) token sequences of fixed length can be used as input within a batch, which speeds up training and b) the model sees distracting background patches in the first input. 
Experiments on masking ratio (ranging from 60\% to 90\%) did not yield any
significant downstream performance difference. We thus use a 75\% masking ratio, in line with~\cite{mae}.

We now detail how to obtain pairs of images of full-bodied human data suitable for pre-training. 
To generate pairs for hands,  we use the InterHand2.6M~\cite{interhand} dataset, from which we sample both cross-view and cross-pose pairs in equal number.

\begin{table*}[tbh]
\newcommand{\video}{\faVideoCamera}
\newcommand{\multiview}{\faCamera}
\centering

\begin{subtable}{0.54\linewidth}

\resizebox{1\textwidth}{!}{%
\begin{tabular}{ccc|cccc}
\toprule
\multicolumn{2}{c|}{multi-view data} & video data & \multicolumn{2}{c}{Body Mesh Recovery} & \multicolumn{2}{c}{DensePose} \\
\multiview & \video & \video & {\small PA-MPJPE$\downarrow$} & {\small MPVPE$\downarrow$} & {\small $AUC_{30}$$\uparrow$} & {\small $AUC_{10}$$\uparrow$} \\
\midrule

\checkmark          &            &                    & 58.6              & 112.3         & 0.655                  & 0.379                  \\ 
                    & \checkmark &                    & 58.1              & 111.8         & 0.645                  & 0.367                  \\ 
\checkmark          & \checkmark &                    & 58.4              & \bf{111.4}    & 0.651                  & 0.372                  \\ 
\hlcell \checkmark  & \hlcell    & \hlcell \checkmark & \hlcell \bf{57.5} & \hlcell 111.5 & \hlcell \textbf{0.692} & \hlcell \textbf{0.416} \\
\bottomrule
\end{tabular}}
\caption{\textbf{Pre-training data for CroCo-body.}
\video \; denotes cross-pose images and \multiview \; indicates cross-view images. }
\end{subtable}
\hfill
\begin{subtable}{0.44\linewidth}
\resizebox{1\textwidth}{!}{%
\renewcommand{\arraystretch}{1.05}
\begin{tabular}{cc|ccc}
\toprule
 \multicolumn{2}{c|}{multi-view data} &  \multicolumn{2}{c}{Hand Mesh Recovery} & Gesture \\
\multiview & \video  & {\small PA-MPJPE$\downarrow$} & {\small MPVPE$\downarrow$} & {\small Acc(\%)$\uparrow$} \\
\midrule

\checkmark         &                                  & \bf{7.97}          & \bf{16.02}    & \bf{97.48}            \\
                   & \checkmark                       & 8.40               & 16.96         & 97.13                \\
\hlcell \checkmark         & \hlcell \checkmark    &                  \hlcell 8.21       & \hlcell 16.69 & \hlcell 97.36 \\
\bottomrule
\end{tabular}

}
\caption{\textbf{Pre-training data for CroCo-Hand.}
\video \; denotes cross-pose images and \multiview \; indicates cross-view images. }
\end{subtable}

\vspace{0.22cm}

\begin{subtable}[t]{0.32\linewidth}

\centering
\small
\setlength{\tabcolsep}{9.2pt}
\resizebox{1\textwidth}{!}{%
\renewcommand{\arraystretch}{1.2}
\setlength{\tabcolsep}{1pt}
\begin{tabular}{lcccc}
\toprule
\multirow{2}{*}{Init.} & \multicolumn{2}{c}{Body Mesh Recovery} & \multicolumn{2}{c}{DensePose} \\
 & {\small PA-MPJPE$\downarrow$} & {\small MPVPE$\downarrow$} & {\small $AUC_{30}$$\uparrow$} & {\small $AUC_{10}$$\uparrow$} \\
\midrule
\hlcell CroCo~\cite{croco} &\hlcell \bf{57.5}  & \hlcell \bf{111.5}   & \hlcell \textbf{0.692} & \hlcell \textbf{0.416} \\
Random & 60.5  & 117.5 & 0.620 & 0.340 \\
\bottomrule
\end{tabular}}
\caption{\textbf{Weights initialization for CroCo-Body.} Starting from pre-trained CroCo weights performs better than a random initialization.}
\end{subtable}  \hfill
\begin{subtable}[t]{0.32\linewidth}
\centering
{
\small
\resizebox{1\textwidth}{!}{%
\renewcommand{\arraystretch}{1.15}
\setlength{\tabcolsep}{1pt}
\begin{tabular}{lcccc}
\toprule
\multirow{2}{*}{Masking} & \multicolumn{2}{c}{Body Mesh Recovery} & \multicolumn{2}{c}{DensePose} \\
 & {\small PA-MPJPE$\downarrow$} & {\small MPVPE$\downarrow$} & {\small $AUC_{30}$$\uparrow$} & {\small $AUC_{10}$$\uparrow$} \\
\midrule
\hlcell Human &\hlcell \bf{57.5}  & \hlcell \bf{111.5}   & \hlcell \textbf{0.692} & \hlcell \textbf{0.416} \\
Uniform & 57.7  & 112.0 & 0.654 & 0.374 \\
\bottomrule
\end{tabular}}

}
\caption{\textbf{Masking strategy.} Performance on downstream tasks improves when the masking is focused on patches where humans are visible.}
\end{subtable}
\hfill
\begin{subtable}[t]{0.32\linewidth}
\centering{
\small
\resizebox{1\textwidth}{!}{%
\setlength{\tabcolsep}{1pt}
\renewcommand{\arraystretch}{1}
\begin{tabular}{lccc}
\toprule
\multirow{2}{*}{Init.} &
\multicolumn{2}{c}{Hand Mesh Recovery} & Gesture \\
 & {\small PA-MPJPE$\downarrow$} & {\small MPVPE$\downarrow$} & {\small Acc(\%)$\uparrow$} \\
\midrule
\hlcell CroCo~\cite{croco} & \hlcell \bf{8.21} & \hlcell \bf{16.69} & \hlcell \bf{97.36}\\
Random & 8.63 & 17.97 & 97.18  \\
\bottomrule
\end{tabular}}
}

\caption{\textbf{Weights initialization for CroCo-Hand.} Hand mesh recovery also benefits from starting from pre-trained CroCo weights.}
\end{subtable} 
\vspace{-0.23cm}
\caption{\textbf{Ablations for \OursBody (left) and \OursHand (right) pre-training strategies}. 
Default settings are highlighted in grey.
}
\vspace{-0.1cm}
\label{tab:data_ab}
\end{table*}

\subsection{Cross-view pair construction} 
\label{subsec:crossview}

While CroCo~\cite{croco} required pairs with rather small camera baselines due to unconstrained scenes, in our case, we work with the strong assumption of a known object -- a person -- and therefore any viewpoint change is admissible.  
 We rely on existing multi-view datasets for two data modalities, namely human \emph{bodies} and \emph{hands}, where subjects are captured from multiple viewpoints by synchronous cameras, selected to have diversity in identities, appearances, poses and backgrounds.
 We use the HUMBI~\cite{humbi} and AIST~\cite{aist} datasets as well as synthetic data.  
 HUMBI
 contains more than $300$ subjects, with a wide range of age, body-shapes, and clothing variety but a restricted set of poses, while AIST sequences are captured from only $40$ subjects with $9$ different camera viewpoints, with plain clothing, but contain a great diversity of poses, with \eg dance moves from professional performers. We gather $1510$ sequences with more than $5M$ images in total.

  \noindent \textbf{Obtaining information for masking.}
  For HUMBI, 
  we run an off-the-shelf human parser, the PGN method from CIHP~\cite{cihp}, to generate silhouette information needed for human masking.
  Note that only a rough estimate of the silhouette is needed for patch-level-guided masking.

  \noindent \textbf{Increased diversity with synthetic data.}
  Taken together, HUMBI and AIST contain a wide variety of identities and poses, but 
  lack diversity in terms of environment, as they
  were captured in an indoor studio setting.
To remedy this, we create a synthetic dataset where we generate multiple-view renderings of SMPL meshes following SURREAL~\cite{Varol_2017_CVPR}, with diverse lighting conditions and camera viewpoint, 
overlaid on top of distracting background images from COCO~\cite{lin2014microsoft} or rendered inside the Habitat simulator~\cite{habitat}.
Body parameters and poses are sampled from AMASS~\cite{Mahmood_2019_ICCV} to generate images with fully-visible persons. 

\subsection{Cross-pose pairs construction}
\label{subsec:crosspose}

As the human body is non-rigid, 
 going beyond the static setting proposed in CroCo~\cite{croco} can enable the model to 
gain some understanding of how body-parts interact and move \wrt one another. 
Dynamic pairs can be 
constructed from a monocular RGB video capturing a human motion.
Such data typically also provide more variety in terms of appearance and background, as they can be captured from a single camera, which is easier to collect in the wild.
We tested various strategies for temporal sampling, using predefined time intervals.
No significant difference was observed on downstream performance. We thus keep a simple random sampling strategy.

\noindent \textbf{Extracting pairs from video datasets.}
We rely on a mix of diverse human-centric datasets namely 3DPW~\cite{3dpw}, Posetrack~\cite{posetrack}, PennAction~\cite{pennaction}, JRDB~\cite{jrdb}, MARS~\cite{mars} and AIST~\cite{aist}.
3DPW includes video footage taken from a moving phone camera of persons performing common actions such as walking or playing guitar. We use videos from the training set only, leading to more than $22k$ frames from $34$ sequences.
Posetrack is a large-scale dataset for multi-person pose estimation and tracking in videos that contains more than $97k$ images from $3719$ sequences.
PennAction contains $2326$ video sequences of 15 different actions.
JRDB contains $310k$ images from $2509$ sequences collected with a social mobile manipulator, JackRabbot.
MARS is a CCTV dataset containing video tracks of more than $1200$ identities and $20k$ video sequences.

\begin{figure*}[tbh]
\begin{subfigure}[b]{0.72\textwidth}
    \vspace*{-3cm} 
    \resizebox{\linewidth}{!}{
    \begin{tabular}{lcccccc}
\toprule
\multirow{2}{*}{Initialization} & \multicolumn{2}{c}{Body Mesh Recovery} & \multicolumn{2}{c}{Hand Mesh Recovery}  & \multicolumn{2}{c}{DensePose} \\
 & PA-MPJPE$\downarrow$ & MPVPE$\downarrow$ & PA-MPJPE$\downarrow$ & MPVPE$\downarrow$ & $AUC_{30}$$\uparrow$ & $AUC_{10}$$\uparrow$\\
\midrule
Random  &  80.2 & 180.0  & 12.92 & 40.06 & 0.430 & 0.206 \\
CroCo~\cite{croco}  & 61.5  & 119.0  & 9.34 & 19.24 & 0.590 & 0.320 \\
MAE-Body/Hand & 59.1 & 115.1 & 8.75 & 17.93 & 0.625 & 0.351 \\
MAE-IN1k~\cite{mae}  & 59.6  & 113.0 & 8.64 & 16.84 & 0.678 & 0.392 \\
\bf{CroCo-Body/Hand}  & \bf{57.4}  & \bf{111.5}  & \bf{7.97} & \bf{16.02} & \textbf{0.692} & \textbf{0.416} \\
\bottomrule
\end{tabular}
     \label{fig:perc_train}
    }
    \begin{center}
    \vspace*{0.05cm}
      {\small (a) Comparison with other pre-training strategies}
    \end{center}
\end{subfigure}
\hfill
\begin{subfigure}[b]{0.25\textwidth}
  \vspace*{-0.5cm}
	\resizebox{1\linewidth}{!}{
    \includegraphics[]{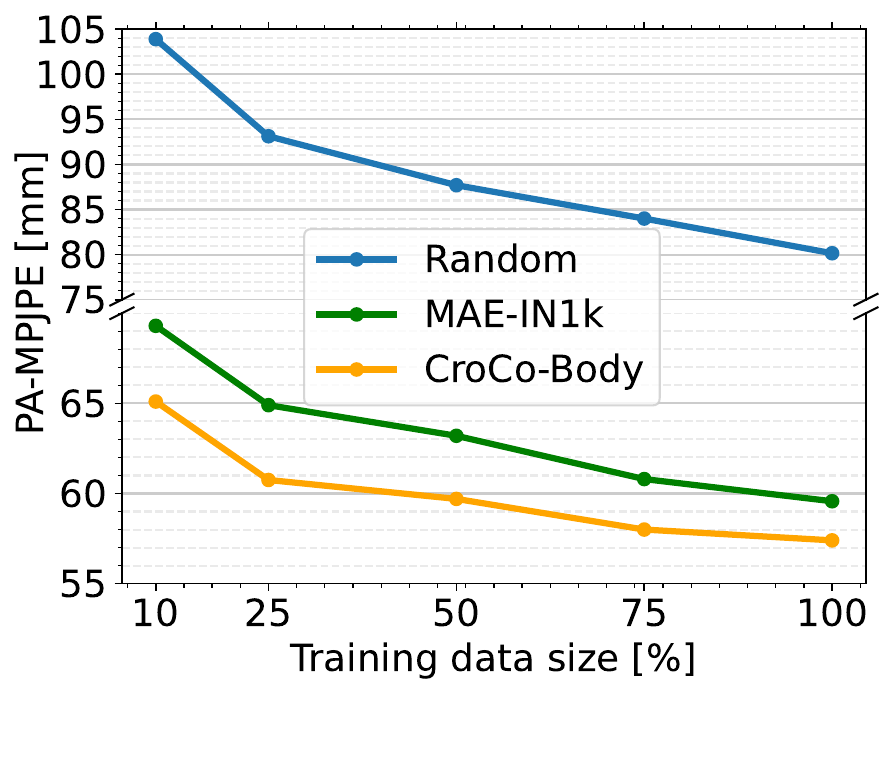}
    }
    \begin{center}
    \vspace*{-0.78cm}
      {\small (b) Data efficiency}
    \end{center}
\end{subfigure}
\vspace*{-0.2cm}
\caption{
    \label{fig:perc_train}
    \textbf{Comparison with other pre-training methods} on different downstream tasks (a) or under different fine-tuning data regimes (b), \ie, when varying the number of annotated training samples from COCO$_{part}$ for fine-tuning on the body mesh recovery task from 10\% to 100\%. MAE-Body/Hand means that we pre-train MAE on the same data as CroCo-Body/Hand.
    }
    \vspace*{-0.4cm}
\end{figure*}

\subsection{Fine-tuning on downstream tasks}
\label{subsec:finetune}

Our model can be fine-tuned on a variety of downstream tasks which can be either image-level (\eg as mesh recovery) or dense, \ie, that require pixelwise predictions (\eg dense vertex coordinates regression). Interestingly, 
our method can tackle monocular and/or binocular tasks.
For monocular task, the ViT-Base encoder is used alone, while binocular tasks benefit from the pre-training of both the encoder and the decoder.
For fine-tuning, we replace the linear head from the pre-training model with a randomly initialized prediction head. This prediction head can be of two types depending on the aforementioned type of task. For image-level tasks, we perform an average pooling on the output tokens from the last transformer block and we use an MLP to regress or classify.
 For dense tasks, we leverage DPT~\cite{dpt}, that assembles tokens from various stages into a dense output with a prediction for each pixel. 

\section{Experiments}
\label{sec:expes}

We evaluate our pre-training strategies on monocular human-centric  tasks listed in Section \ref{sec:tasks}. 
We then provide extensive ablations (Section \ref{sec:ablations}) and comparison with prior works (Section \ref{sec:sota}).
Finally we show results in binocular settings in Section~\ref{sec:binocular}.
We provide qualitative examples for several downstream tasks in Appendix~\ref{sec:vizu}.

\subsection{Downstream tasks}
\label{sec:tasks}

\noindent \textbf{Body mesh recovery.}
We evaluate \OursBody on the body mesh recovery task on the 3DPW test set \cite{3dpw} composed of 35k human crops following prior works~\cite{kanazawa2018end,joo2021exemplar,kolotouros2019learning}.
We use different type of training data for fine-tuning: a) \texttt{COCO-Part-EFT}~\cite{joo2021exemplar} which consists of 28k images from COCO with pseudo-ground-truth of SMPL parameters \cite{loper2015smpl}, b) \texttt{Mix}: a mix of 2D/3D datasets commonly used by prior works~\cite{kanazawa2018end,joo2021exemplar,kolotouros2019learning} composed of COCO-All-EFT~\cite{joo2021exemplar}, MPII~\cite{andriluka14cvpr}, Human3.6M~\cite{h36m_pami} and MPI-INF-3DHP~\cite{mono-3dhp2017}), and c) \texttt{Mix$^*$}: the mix of datasets described in b) plus 3DPW train set when comparing against the state of the art.
When not specified, we use \texttt{COCO-Part-EFT} for fine-tuning on this task. 
a) a model-based method using a vanilla `iterative regressor' \cite{kanazawa2018end,joo2021exemplar,kolotouros2019learning} prediction head, which consists in regressing the SMPL parameters with an MLP in an iterative manner starting from the mean parameters, and b) a model-free method which directly regresses the 3D location of the vertices using the METRO prediction head~\cite{lin2021end}.
We report standard metrics: MPJPE (Mean Per Joint Error), MPVPE (Mean Per Vertex Error) and PA-MPJPE (Procrustes-Aligned MPJPE) in millimeters.

\noindent \textbf{DensePose.}
We also evaluate on the DensePose task~\cite{guler2018densepose}, which consists in mapping every human pixel in an image to a point on the surface of the body.
It involves dense classification (of body parts) and dense regression (UV coordinates).
Our output and losses are similar to DensePose: we learn a 25-way classification unit, and 24 2D regression functions. 
We do not target detection, so we train and evaluate on single-person human-centered crops only.
For evaluation, we compute the ratio of correct points $RCP(t)$ among annotated pixels, where a correspondence is considered correct if the geodesic distance (on the surface) between estimated pixel value and annotation is below a threshold $t$. Next, taking all annotated points on the evaluation set, we consider the area under the curve (AUC), defined as $AUC_a = \frac{1}{a} \int_{0}^{a} RCP(t) dt$, for $a=30$ and $a=10$.

\noindent\textbf{Hand mesh recovery.}
We fine-tune \OursHand on the task of hand mesh recovery.
The setup is similar to the body mesh recovery task with the MANO parametric model~\cite{romero2022embodied}.
We only use the vanilla `iterative regressor' head~\cite{kanazawa2018end,joo2021exemplar,kolotouros2019learning} and we report results on the FreiHand dataset~\cite{zimmermann2019freihand}. 

\noindent \textbf{Hand gesture classification.}
 We fine-tune \OursHand on HaGRID~\cite{hagrid}, which contains 18 gesture classes with 509,323 images for training and 43,669 for testing.
Using the full training set typically leads to very high accuracies (\eg above 99\%), likely thanks to the large amount of annotated samples, and in this regime, unsupervised pre-training is thus unnecessary. In the regime where little data is available, however, pre-training is expected to have a noticeable impact. We thus report few-shot accuracy considering 64 samples per class except otherwise stated.

\noindent \textbf{Grasp classification.} 
We fine-tune \OursHand on the Grasp UNderstanding (GUN-71) dataset~\cite{rogez2015understanding}, which consists of 12k first-person RGB images annotated with 71 hand-object manipulation classes.
We follow the evaluation protocol of~\cite{rogez2015understanding} on the 33 grasp taxonomy defined in~\cite{Feix_33grasps_2009} with the `Best View' and `All' settings.
This dataset is of particular interest as the classes are fine-grained, with some highly-specific grasps.

\begin{figure*}[!tbh]
\begin{minipage}{0.59\linewidth}
\includegraphics[width=\linewidth]{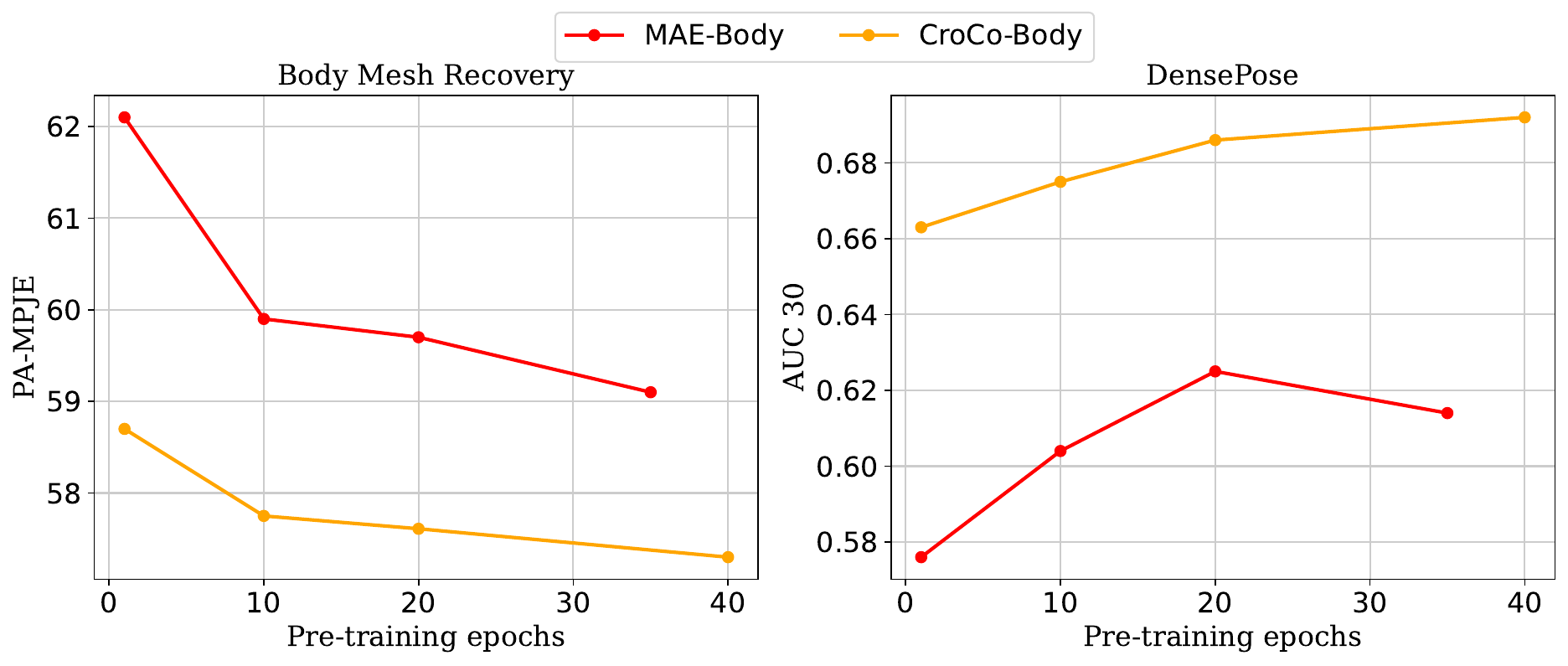} \\[-0.8cm]
\caption{
\label{fig:masking_ratio}
\textbf{Impact of the number of pre-training epochs.} 
\OursBody is initialized from CroCo while MAE is initialized from ImageNet. 
}
\end{minipage}
\hfill
\begin{minipage}{0.39\linewidth}
\centering
\vspace*{0.7cm}
\resizebox{\linewidth}{!}{
\begin{tabular}{lc@{~~}c@{~~}cc@{~~}c}
\toprule
\multirow{2}{*}{Initialization} & \multicolumn{3}{c}{Gesture$\uparrow$} & \multicolumn{2}{c}{Grasp$\uparrow$} \\
& 16 & 32 & 64 & BV & All \\ 
\midrule
Random & 45.7 & 68.4 & 81.2 & 10.9 & ~8.6 \\
MAE-Hand &  85.1 & 92.7 & 95.9 & 33.2 & 25.8 \\
\bf{\OursHand} & \bf{92.5}  & \bf{95.8} & \bf{97.4} & \bf{37.8} & \bf{29.2} \\
\bottomrule
\end{tabular}
}
\vspace{-0.3cm}
\captionof{table}{
\label{tab:hagrid}
\textbf{Impact of pre-training method for gesture and grasp classification.} For gesture, we report the accuracy (\%) when varying the number of samples per class used for training. For grasp, we report both the `Best View' (BV) and `All' protocols from~\cite{rogez2015understanding}.
}
\end{minipage}
\vspace{-0.1cm}
\end{figure*}

\subsection{Ablation studies}
\label{sec:ablations}

\noindent \textbf{Pair construction.} In Table~\ref{tab:data_ab} (a) and (b), we construct cross-view and cross-pose pairs only from datasets that can be used to create \emph{both} types of pairs -- \ie, for body-centric data, multi-view datasets that also include some temporal sampling of motion, namely HUMBI, AIST and synthetic data, InterHand for hand-centric data. This way, the pair construction method is the only difference and the data used is the same across variants and results are not influenced by the use of additional data.
Overall, the best results are obtained with cross-view pairs,
and it seems they should be favored when possible. However, the diversity of multi-view data available is limited, and enriching the training data mix with cross-pose pairs from monocular videos leads to a clear performance boost, see last row of Table~\ref{tab:data_ab} (a).

\noindent \textbf{Initialization.} In Table~\ref{tab:data_ab} (c) and (e) we start the training of CroCo-Body and CroCo-Hand from random weights, instead of weights initialized from CroCo pre-training (default setting), and train for the same number of iterations. We observe that this decreases the results on all tasks.

\noindent \textbf{Masking.}  In Table~\ref{tab:data_ab} (d), we investigate the importance of the masking strategy in our pre-training.
We compare our human-centric masking strategy 
to a naive \emph{uniform} masking strategy which consists in randomly sampling tokens from the image. 
Performance degrades across tasks when using the uniform sampling: we conclude that masking pixels that belong to the person with higher probability is beneficial to learning human-centric features.

\noindent \textbf{Comparison with other pre-training strategies.}
In Figure~\ref{fig:perc_train} (a), we compare our pre-trained \OursBody and \OursHand models with the publicly released CroCo~\cite{croco} and MAE~\cite{mae} weights.
In order to evaluate the respective contributions of pre-training data and pre-training method, we also pre-train MAE on our data.
We refer to these models as MAE-Body/MAE-Hand and refer to the original MAE model as MAE-IN1k.
Our pre-training (bottom row) outperforms both the CroCo and MAE-IN1k models. 
Interestingly, MAE-Body/Hand performs worse than MAE-IN1k, thus the gains of our models come from the pre-training objective rather than from the difference in pre-training data. This is partly unexpected, as our pre-training data is more tailored to the downstream tasks, 
but may be explained by the fact 
that ImageNet-1k offers a vast
diversity of backgrounds and texture, which may be beneficial for pre-training. 
Figure~\ref{fig:masking_ratio} shows the evolution of performance on downstream tasks after a varying number of epochs; it shows that CroCo-Body starts off from a better performance, and converges fast compared to MAE-Body.

In Table~\ref{tab:hagrid}, we compare our \OursHand 
to MAE-Hand and a random initialization on the hand classification tasks (gesture and grasp). In all cases, we 
outperform MAE-Hand, which itself outperforms a random initialization.
 
\begin{table}
\centering
{
\small
\begin{tabular}{lcc}
\toprule
\multirow{2}{*}{Pre-training supervision} & \multicolumn{2}{c}{Body Mesh Recovery} \\
  & PA-MPJPE $\downarrow$ & MPVPE $\downarrow$ \\
\hline
2D keypoints (supervised) & 54.6  & 116.7  \\
\OursBody (self-supervised) & 53.8  & 112.8 \\
\OursBody + 2D keypoints & \bf{52.8}  & \bf{111.5}  \\
\bottomrule
\end{tabular}
}
\vspace{-0.3cm}
\caption{\textbf{Ablation on pretext task used for pre-training}. 
All models are pre-trained only on datasets for which we have keypoints annotations, namely 3DPW, PoseTrack, PennAction, AIST, and JRDB.
We use \texttt{Mix} for fine-tuning.
}
\vspace{-0.3cm}
\label{tab:kpts_ab}
\end{table}

\noindent \textbf{Fine-tuning data efficiency.}
One motivation for pre-training is that the fine-tuning stage on a specific task requires less annotated training samples to reach good performance.
In Figure~\ref{fig:perc_train} (b), we report the performance on body tasks of different pre-training strategies and their efficiency under different fine-tuning data regimes. 
\OursBody achieves better performance compared to other pre-training strategies (MAE-IN1K and Random) with the same number of training samples: it reaches the same performance as MAE-Body with 2 to 3 times less training samples.
In particular, \OursBody obtains a PA-MPJPE of 65mm with 10\% of the training set while MAE-IN1K requires 25\% of the train set to reach a similar performance.
Table~\ref{tab:hagrid} reports the results for grasp recognition and for few-shot gesture classification with 16, 32 and 64 samples per class  for a network initialized randomly, with MAE-Hand or with \OursHand.
\OursHand outperforms MAE-Hand which itself obtains significantly higher accuracies than a random initialization. 
Interestingly, \OursHand obtains above 92\% accuracy even with only 16 shots, while MAE-Hand requires 32 samples per class to reach such accuracy.

\noindent \textbf{Keypoints supervision. }
We evaluate how the proposed self-supervised pre-training strategy compares to a fully-supervised pre-training.
2D keypoints annotations are available for multiple human-centric datasets, and we thus pre-train a network in a fully-supervised manner to regress body keypoints as dense 2D heatmaps, following usual practice in the field~\cite{sun2021monocular,cho2022FastMETRO}.
We fine-tune this model on human-related downstream tasks, and compare performance with the proposed \OursBody pre-training on the same data.
\OursBody pre-training led to better performances than the supervised approach for all downstream tasks considered (see Table~\ref{tab:kpts_ab}), confirming the interest of the proposed approach. Yet, by combining 2D keypoints supervision to cross-view completion during pre-training, we were able to obtain a small additional performance boost.

\begin{table}[t]
    \begin{subtable}{\linewidth}
    \centering
     \setlength{\tabcolsep}{10pt}
     \resizebox{\linewidth}{!}{
    \begin{tabular}{l@{~~~~}l@{~~~~~~}c@{~~}c}      
        \toprule
         & Method & PA-MPJPE$\downarrow$ & MPJPE$\downarrow$ \\
         \midrule
        {\multirow{5}{*}{\rotatebox[origin=c]{90}{model-based}}}  &  HybrIK~\cite{li2021hybrik} & 48.8 & 80.0 \\ 
        & ROMP~\cite{sun2021monocular} & 47.3 & 76.7 \\
        & PARE~\cite{kocabas2021pare} & \underline{46.5} & \bf{74.5} \\
        & Choi \etal \cite{choi2023rethinking} & 56.8 & - \\
        & \bf{\OursBody} & \bf{46.2} & \underline{76.2} \\
        \midrule
        {\multirow{6}{*}{\rotatebox[origin=c]{90}{model-free}}} & Pose2Mesh~\cite{choi2020pose2mesh} & 56.3 & 89.5 \\
        & METRO~\cite{lin2021end} & 47.9 & 77.1 \\
        & MeshGraphormer~\cite{lin2021-mesh-graphormer} & 45.6 & 74.7 \\
        & FastMETRO~\cite{cho2022FastMETRO} & \underline{44.6} & \underline{73.5} \\
        & PointHMR~\cite{PointHMR} & 44.9 & 73.9 \\
        & \bf{\OursBody} & \bf{44.2} & \bf{73.3} \\
        \bottomrule
    \end{tabular}
    }
    \vspace{-0.1cm}
    \caption{Body mesh recovery on 3DPW~\cite{3dpw} after fine-tuning on 
 \texttt{Mix*}}
    \end{subtable}
\begin{subtable}{\linewidth}
\vspace{0.2cm}
    \centering
    \renewcommand{\arraystretch}{1.}
    \setlength{\tabcolsep}{10pt}
    \resizebox{\linewidth}{!}{
    \begin{tabular}{l@{\hskip 1.55cm}cc}
      \toprule
      Method & AUC$_{10}\uparrow$ & AUC$_{30}\uparrow$ \\ 
      \midrule
        DensePose~\cite{guler2018densepose} & 0.378 & 0.614 \\
        \bf{\OursBody} & \textbf{0.416} & \textbf{0.692} \\ 
        \bottomrule
      \end{tabular}
    }
    \vspace{-0.1cm}
    \caption{DensePose estimation on COCO~\cite{guler2018densepose}}
    \end{subtable}
\vspace{-0.7cm}
\caption{\textbf{\OursBody \vs the state of the art}. Using a generic architecture, we achieve competitive performances with recent works thanks to the proposed pre-training.
}
\vspace*{-0.2cm}
\label{tab:sota}
\end{table}

\begin{table}
 \begin{subtable}{\linewidth}
    \centering
    \vspace{0.2cm}
    \resizebox{\linewidth}{!}{
    \begin{tabular}{lcc}
    \toprule
      Method & PA-MPJPE$\downarrow$ & PA-MPVPE$\downarrow$ \\ \hline
         PIXIE {\small hand expert}~\cite{feng2021collaborative} & 12.0 & 12.1 \\
         MANO-CNN~\cite{zimmermann2019freihand} &  11.0 & 10.9 \\
         Pose2Mesh~\cite{choi2020pose2mesh} &  7.7 & 7.8 \\
         Hand4Whole~\cite{moon2022accurate} &  7.7 & \underline{7.7} \\
         METRO~\cite{lin2021end} &  \textbf{6.8} & \textbf{6.7} \\
         \bf{\OursHand} & \underline{7.4} & 8.0 \\
    \bottomrule
    \end{tabular}
    }
    \vspace{-0.1cm}
    \caption{Hand mesh recovery on FreiHand~\cite{zimmermann2019freihand}}
    \end{subtable}
\begin{subtable}[t]{\linewidth}
\centering
\resizebox{\linewidth}{!}{
\begin{tabular}{l@{\hskip 1.8cm}ccc|cc|cc}
\toprule
Method & Best View$\uparrow$ & All$\uparrow$ \\ \midrule 
Rogez \etal~\cite{rogez2015understanding} &  22.7 & 20.5  \\
Coskun \etal~\cite{coskun2021domain} &  \underline{42.3} & N/A  \\
Choi \textit{et al.}~\cite{robusthand} & N/A & \underline{39.8}   \\
\bf{\OursHand} & 37.8 & 29.2  \\
\bf{\OursHand$\!^{\dagger}$} & \bf{49.9} & \bf{49.6}  \\
\bottomrule
\end{tabular}
}
\vspace{-0.1cm}
\caption{Grasp classification on GUN-71~\cite{rogez2015understanding}}
\end{subtable}

    \vspace{-0.2cm}
    \caption{\textbf{\OursHand \vs the state of the art.} \OursHand$\!^{\dagger}$ represents our method pre-trained with more diverse data.
    }
    \vspace*{-0.2cm}
    \label{tab:sotahand}
\end{table}

\subsection{Comparison to the state of the art}
\label{sec:sota}

We now
compare our performance against state-of-the-art methods, reported in Table \ref{tab:sota} and \ref{tab:sotahand}.

On the task of Body Mesh Recovery (Table \ref{tab:sota} (a)), \OursBody achieves state-of-the-art performances using either a model-based or model-free head.
\OursBody yields consistent gains across choices of head, which indicates that our proposed pre-training stage learns highly transferable features for the task of mesh recovery.

We report in Table \ref{tab:sota} (b) the results on the DensePose task. We follow the setup of \cite{guler2018densepose} and observe that \OursBody achieves better performance on all metrics.

Next, we report the results on hand mesh recovery in Table \ref{tab:sotahand} (a). \OursHand gets results on par with prior work.
Recent methods leverage more complicated prediction heads (see Section~\ref{sec:tasks}), however, we show that without bells and whistles, we perform on par with these methods.

Finally, we report grasp classification in Table~\ref{tab:sotahand} (b). \OursHand is competitive with prior works~\cite{rogez2015understanding,coskun2021domain,robusthand}  but does not reach state-of-the-art performances.
This can be explained by the extremely limited variety of backgrounds and the absence of manipulated objects in the pre-training dataset~\cite{interhand}. 
Pre-training on diverse images seems to be critical for reaching high performance on GUN-71, as suggested by the results of Coskun \etal~\cite{coskun2021domain} who pre-trained their network on ImageNet.
To solve this issue, we pre-train \OursHand using cross-view and cross-pose pairs from more datasets containing hand-object interactions and more diverse backgrounds (InterHand2.6M~\cite{interhand}, DexYCB~\cite{chao2021dexycb}, HO3D~\cite{hampali2020honnotate}, HanCo~\cite{hanco}). This variant, denoted as \OursHand$\!^{\dagger}$, leads to a clear improvement, reaching state-of-the-art accuracy.

\subsection{Extension to binocular tasks}
\label{sec:binocular}
                   
One interesting property of our \OursBody and \OursHand architecture is that the decoder can also be leveraged in binocular settings.
In Table~\ref{tab:mono_bino_body_mesh_recovery} we report performance of \OursBody and \OursHand using either a single-image input with the encoder only, or using two inputs images and leveraging both the encoder and decoder.
For the body mesh recovery task, we fine-tune on COCO-Part and the 3DPW train set as well since 3DPW contains sequences, so we can feed  neighboring frames to our model as well.
For hand mesh recovery, we fine-tune on Hanco~\cite{hanco}, a multi-view extension of FreiHand. 
Results for the texture estimation task are reported in Appendix~\ref{sec:texture}.

Using both the pre-trained encoder and decoder for fine-tuning in a binocular framework proves to be beneficial for both body and hand mesh recovery. 
For sanity’s sake, we also fine-tune MAE-IN1k in binocular settings by initializing the decoder from scratch. While it degrades the performance on the body mesh recovery task compared to its monocular counterpart, we observe that it gets reasonable performance on the hand mesh recovery task.
This can be explained by the fact that the HanCo training set is large compared to the training set used for the body mesh recovery task so the pre-training has less of an impact. 
\OursBody/Hand 
still obtains better performance than MAE-IN1k in both monocular and binocular settings.

\begin{table}
    \centering
    \resizebox{\linewidth}{!}{
    \setlength{\tabcolsep}{4pt}
    \small
\begin{tabular}{cccccc}
    \toprule
        \multirow{2}{*}{Input} & 
        \multirow{2}{*}{Pre-training} & \multicolumn{2}{c}{Hand Mesh Recovery} & \multicolumn{2}{c}{Body Mesh Recovery} \\
                                   &                              & {\scriptsize PA-MPJPE}$\downarrow$                     & {\scriptsize PA-MPVPE}$\downarrow$ & {\scriptsize PA-MPJPE}$\downarrow$                      & {\scriptsize PA-MPVPE}$\downarrow$ \\ 
       \midrule
        \multirow{2}{*}{Monocular}                           & MAE-IN1k                     & 11.6                            	      & 11.2                  & 58.9                                       & 113.0     \\            
                                   & \bf{CroCo-Body/Hand}              & \bf{11.0}                                      & \bf{10.8}  		      & \bf{57.5}                                       & \bf{111.2}     \\           
     \midrule                                \multirow{2}{*}{Binocular}                                                                                                                                                                      
                         & MAE-IN1k                    & 9.9                                       & 9.5                   & 60.5                                       & 115.0     \\          
                         & \bf{CroCo-Body/Hand}             & \bf{9.5}                                  & \bf{8.8}              & \bf{55.0}                                  & \bf{108.4}\\ 
         \bottomrule
    \end{tabular}

    } \vspace{-0.3cm}
    \caption{\textbf{Monocular \vs binocular mesh recovery} performance on the 3DPW test set (body) and on HanCo (hand). 
    }
    \vspace{-0.3cm}
    \label{tab:mono_bino_body_mesh_recovery}
\end{table}

\section{Conclusion}

 We present a strategy to pre-train a model for various human-centric tasks. Our approach is based on a self-supervised pre-training method that leverages cross-view completion (CroCo), using pairs of images of people taken from different viewpoints or in different poses.
 Fine-tuning a model pre-trained with this strategy leads to state-of-the-art performance on various tasks, such as human mesh recovery and competitive performance on gesture classification, while using a generic model architecture. The proposed pre-training outperforms other popular pre-training strategies such as MAE~\cite{mae}. 
 Lastly, we show how this pre-training can be leveraged to perform binocular tasks such as body mesh recovery from multiple views. 

{
\bibliographystyle{ieee_fullname}
\bibliography{biblio}

\begin{thebibliography}{10}\itemsep=-1pt

\bibitem{andriluka14cvpr}
Mykhaylo Andriluka, Leonid Pishchulin, Peter Gehler, and Bernt Schiele.
\newblock 2d human pose estimation: New benchmark and state of the art
  analysis.
\newblock In {\em CVPR}, 2014.

\bibitem{assran2022hidden}
Mahmoud Assran, Randall Balestriero, Quentin Duval, Florian Bordes, Ishan
\newblock The hidden uniform cluster prior in self-supervised learning.
\newblock In {\em ICLR}, 2023.

\bibitem{msn}
Mahmoud Assran, Mathilde Caron, Ishan Misra, Piotr Bojanowski, Florian Bordes,
  Pascal Vincent, Armand Joulin, Michael Rabbat, and Nicolas Ballas.
\newblock Masked siamese networks for label-efficient learning.
\newblock In {\em ECCV}, 2022.

\bibitem{sit}
Sara Atito, Muhammad Awais, and Josef Kittler.
\newblock {SiT: Self-supervised vIsion Transformer}.
\newblock {\em arXiv preprint arXiv:2104.03602}, 2021.

\bibitem{multimae}
Roman Bachmann, David Mizrahi, Andrei Atanov, and Amir Zamir.
\newblock {MultiMAE: Multi-modal Multi-task Masked Autoencoders}.
\newblock In {\em ECCV}, 2022.

\bibitem{data2vec}
Alexei Baevski, Wei-Ning Hsu, Qiantong Xu, Arun Babu, Jiatao Gu, and Michael
  Auli.
\newblock {data2vec: A General Framework for Self-supervised Learning in
  Speech, Vision and Language}.
\newblock {\em arXiv preprint arXiv:2202.03555}, 2022.

\bibitem{brown2020language}
Tom Brown, Benjamin Mann, Nick Ryder, Melanie Subbiah, Jared~D Kaplan, Prafulla
  Dhariwal, Arvind Neelakantan, Pranav Shyam, Girish Sastry, Amanda Askell,
  et~al.
\newblock Language models are few-shot learners.
\newblock In {\em NeurIPS}, 2020.

\bibitem{caron2018deep}
Mathilde Caron, Piotr Bojanowski, Armand Joulin, and Matthijs Douze.
\newblock Deep clustering for unsupervised learning of visual features.
\newblock In {\em ECCV}, 2018.

\bibitem{swav}
Mathilde Caron, Ishan Misra, Julien Mairal, Priya Goyal, Piotr Bojanowski, and
  Armand Joulin.
\newblock {Unsupervised Learning of Visual Features by Contrasting Cluster
  Assignments}.
\newblock In {\em {NeurIPS}}, 2020.

\bibitem{dino}
Mathilde Caron, Hugo Touvron, Ishan Misra, Herv\'{e} J\'{e}gou, Julien Mairal,
  Piotr Bojanowski, and Armand Joulin.
\newblock {Emerging Properties in Self-Supervised Vision Transformers}.
\newblock In {\em ICCV}, 2021.

\bibitem{carreira2019short}
Joao Carreira, Eric Noland, Chloe Hillier, and Andrew Zisserman.
\newblock A short note on the kinetics-700 human action dataset.
\newblock {\em arXiv preprint arXiv:1907.06987}, 2019.

\bibitem{chao2021dexycb}
Yu-Wei Chao, Wei Yang, Yu Xiang, Pavlo Molchanov, Ankur Handa, Jonathan
  Tremblay, Yashraj~S. Narang, Karl~Van Wyk, Umar Iqbal, Stan Birchfield, Jan
  Kautz, and Dieter Fox.
\newblock Dexycb: A benchmark for capturing hand grasping of objects.
\newblock In {\em CVPR}, 2021.

\bibitem{iGPT}
Mark Chen, Alec Radford, Rewon Child, Jeffrey Wu, Heewoo Jun, David Luan, and
  Ilya Sutskever.
\newblock {Generative Pretraining From Pixels}.
\newblock In {\em {ICML}}, 2020.

\bibitem{simclr}
Ting Chen, Simon Kornblith, Mohammad Norouzi, and Geoffrey Hinton.
\newblock A simple framework for contrastive learning of visual
  representations.
\newblock In {\em {ICML}}, 2020.

\bibitem{chen2023beyond}
Weihua Chen, Xianzhe Xu, Jian Jia, Hao Luo, Yaohua Wang, Fan Wang, Rong Jin,
  and Xiuyu Sun.
\newblock Beyond appearance: a semantic controllable self-supervised learning
  framework for human-centric visual tasks.
\newblock In {\em CVPR}, 2023.

\bibitem{liftedcl}
Ziwei Chen, Qiang Li, Xiaofeng Wang, and Wankou Yang.
\newblock Liftedcl: Lifting contrastive learning for human-centric perception.
\newblock In {\em ICLR}, 2023.

\bibitem{genebody}
Wei Cheng, Su Xu, Jingtan Piao, Chen Qian, Wayne Wu, Kwan-Yee Lin, and
  Hongsheng Li.
\newblock Generalizable neural performer: Learning robust radiance fields for
  human novel view synthesis.
\newblock {\em arXiv preprint arXiv:2204.11798}, 2022.

\bibitem{cho2022FastMETRO}
Junhyeong Cho, Kim Youwang, and Tae-Hyun Oh.
\newblock Cross-attention of disentangled modalities for 3d human mesh recovery
  with transformers.
\newblock In {\em ECCV}, 2022.

\bibitem{robusthand}
Chiho Choi, Sang~Ho Yoon, Chin-Ning Chen, and Karthik Ramani.
\newblock Robust hand pose estimation during the interaction with an unknown
  object.
\newblock In {\em ICCV}, 2017.

\bibitem{choi2020pose2mesh}
Hongsuk Choi, Gyeongsik Moon, and Kyoung~Mu Lee.
\newblock Pose2mesh: Graph convolutional network for 3d human pose and mesh
  recovery from a 2d human pose.
\newblock In {\em ECCV}, 2020.

\bibitem{choi2023rethinking}
Hongsuk Choi, Hyeongjin Nam, Taeryung Lee, Gyeongsik Moon, and Kyoung~Mu Lee.
\newblock Rethinking self-supervised visual representation learning in
  pre-training for 3d human pose and shape estimation.
\newblock {\em ICLR}, 2023.

\bibitem{ci2023unihcp}
Yuanzheng Ci, Yizhou Wang, Meilin Chen, Shixiang Tang, Lei Bai, Feng Zhu, Rui
  Zhao, Fengwei Yu, Donglian Qi, and Wanli Ouyang.
\newblock Unihcp: A unified model for human-centric perceptions.
\newblock In {\em CVPR}, 2023.

\bibitem{coskun2021domain}
Huseyin Coskun, M~Zeeshan Zia, Bugra Tekin, Federica Bogo, Nassir Navab,
  Federico Tombari, and Harpreet Sawhney.
\newblock Domain-specific priors and meta learning for few-shot first-person
  action recognition.
\newblock {\em IEEE Trans. PAMI}, 2021.

\bibitem{deng2009imagenet}
Jia Deng, Wei Dong, Richard Socher, Li-Jia Li, Kai Li, and Li Fei-Fei.
\newblock Imagenet: A large-scale hierarchical image database.
\newblock In {\em CVPR}, 2009.

\bibitem{BERT}
Jacob Devlin, Ming-Wei Chang, Kenton Lee, and Kristina Toutanova.
\newblock {BERT: Pre-training of Deep Bidirectional Transformers for Language
  Understanding}.
\newblock In {\em {NAACL HLT}}, 2019.

\bibitem{splitmask}
Alaaeldin El-Nouby, Gautier Izacard, Hugo Touvron, Ivan Laptev, Herv\'{e}
  Jegou, and Edouard Grave.
\newblock {Are Large-scale Datasets Necessary for Self-Supervised
  Pre-training?}
\newblock {\em arXiv preprint arXiv:2112.10740}, 2021.

\bibitem{spatioMae}
Christoph Feichtenhofer, Haoqi Fan, Yanghao Li, and Kaiming He.
\newblock Masked autoencoders as spatiotemporal learners.
\newblock In {\em NeurIPS}, 2022.

\bibitem{Feix_33grasps_2009}
T. Feix, R. Pawlik, H. Schmiedmayer, J. Romero, and D. Kragic.
\newblock A comprehensive grasp taxonomy.
\newblock In {\em Robotics, Science and Systems: Workshop on Understanding the
  Human Hand for Advancing Robotic Manipulation}, 2009.

\bibitem{feng2021collaborative}
Yao Feng, Vasileios Choutas, Timo Bolkart, Dimitrios Tzionas, and Michael~J
  Black.
\newblock Collaborative regression of expressive bodies using moderation.
\newblock In {\em 3DV}, 2021.

\bibitem{cihp}
Ke Gong, Xiaodan Liang, Yicheng Li, Yimin Chen, Ming Yang, and Liang Lin.
\newblock Instance-level human parsing via part grouping network.
\newblock In {\em ECCV}, 2018.

\bibitem{byol}
Jean{-}Bastien Grill, Florian Strub, Florent Altch{\'{e}}, Corentin Tallec,
  Pierre~H. Richemond, Elena Buchatskaya, Carl Doersch, Bernardo~{\'{A}}vila
  Pires, Zhaohan Guo, Mohammad~Gheshlaghi Azar, Bilal Piot, Koray Kavukcuoglu,
  R{\'{e}}mi Munos, and Michal Valko.
\newblock {Bootstrap Your Own Latent - {A} New Approach to Self-Supervised
  Learning}.
\newblock In {\em {NeurIPS}}, 2020.

\bibitem{guler2018densepose}
R{\i}za~Alp G{\"u}ler, Natalia Neverova, and Iasonas Kokkinos.
\newblock Densepose: Dense human pose estimation in the wild.
\newblock In {\em CVPR}, 2018.

\bibitem{hampali2020honnotate}
Shreyas Hampali, Mahdi Rad, Markus Oberweger, and Vincent Lepetit.
\newblock Honnotate: A method for 3d annotation of hand and object poses.
\newblock In {\em CVPR}, 2020.

\bibitem{mae}
Kaiming He, Xinlei Chen, Saining Xie, Yanghao Li, Piotr Doll\'{a}r, and Ross
  Girshick.
\newblock {Masked Autoencoders are Scalable Vision Learners}.
\newblock In {\em CVPR}, 2022.

\bibitem{hong2022hcmoco}
Fangzhou Hong, Liang Pan, Zhongang Cai, and Ziwei Liu.
\newblock Versatile multi-modal pre-training for human-centric perception.
\newblock In {\em CVPR}, 2022.

\bibitem{h36m_pami}
Catalin Ionescu, Dragos Papava, Vlad Olaru, and Cristian Sminchisescu.
\newblock Human3. 6m: Large scale datasets and predictive methods for 3d human
  sensing in natural environments.
\newblock {\em IEEE Trans. PAMI}, 2013.

\bibitem{posetrack}
Umar Iqbal, Anton Milan, and Juergen Gall.
\newblock Posetrack: Joint multi-person pose estimation and tracking.
\newblock In {\em CVPR}, 2017.

\bibitem{jakab2018unsupervised}
Tomas Jakab, Ankush Gupta, Hakan Bilen, and Andrea Vedaldi.
\newblock Unsupervised learning of object landmarks through conditional image
  generation.
\newblock {\em NeurIPS}, 2018.

\bibitem{jakab2020self}
Tomas Jakab, Ankush Gupta, Hakan Bilen, and Andrea Vedaldi.
\newblock Self-supervised learning of interpretable keypoints from unlabelled
  videos.
\newblock In {\em CVPR}, 2020.

\bibitem{joo2021exemplar}
Hanbyul Joo, Natalia Neverova, and Andrea Vedaldi.
\newblock Exemplar fine-tuning for 3d human model fitting towards in-the-wild
  3d human pose estimation.
\newblock In {\em 3DV}, 2021.

\bibitem{kanazawa2018end}
Angjoo Kanazawa, Michael~J Black, David~W Jacobs, and Jitendra Malik.
\newblock End-to-end recovery of human shape and pose.
\newblock In {\em CVPR}, 2018.

\bibitem{hagrid}
Alexander Kapitanov, Andrey Makhlyarchuk, and Karina Kvanchiani.
\newblock Hagrid - hand gesture recognition image dataset.
\newblock {\em arXiv preprint arXiv:2206.08219}, 2022.

\bibitem{PointHMR}
Jeonghwan Kim, Mi-Gyeong Gwon, Hyunwoo Park, Hyukmin Kwon, Gi-Mun Um, and
  Wonjun Kim.
\newblock {Sampling is Matter}: Point-guided 3d human mesh reconstruction.
\newblock In {\em CVPR}, 2023.

\bibitem{kocabas2021pare}
Muhammed Kocabas, Chun-Hao~P Huang, Otmar Hilliges, and Michael~J Black.
\newblock Pare: Part attention regressor for 3d human body estimation.
\newblock In {\em CVPR}, 2021.

\bibitem{kolotouros2019learning}
Nikos Kolotouros, Georgios Pavlakos, Michael~J Black, and Kostas Daniilidis.
\newblock Learning to reconstruct 3d human pose and shape via model-fitting in
  the loop.
\newblock In {\em ICCV}, 2019.

\bibitem{krishna2017visual}
Ranjay Krishna, Yuke Zhu, Oliver Groth, Justin Johnson, Kenji Hata, Joshua
  Kravitz, Stephanie Chen, Yannis Kalantidis, Li-Jia Li, David~A Shamma, et~al.
\newblock Visual genome: Connecting language and vision using crowdsourced
  dense image annotations.
\newblock {\em IJCV}, 2017.

\bibitem{li2021hybrik}
Jiefeng Li, Chao Xu, Zhicun Chen, Siyuan Bian, Lixin Yang, and Cewu Lu.
\newblock Hybrik: A hybrid analytical-neural inverse kinematics solution for 3d
  human pose and shape estimation.
\newblock In {\em CVPR}, 2021.

\bibitem{mannequinchallenge}
Zhengqi Li, Tali Dekel, Forrester Cole, Richard Tucker, Noah Snavely, Ce Liu,
  and William~T Freeman.
\newblock Learning the depths of moving people by watching frozen people.
\newblock In {\em CVPR}, 2019.

\bibitem{lin2021end}
Kevin Lin, Lijuan Wang, and Zicheng Liu.
\newblock End-to-end human pose and mesh reconstruction with transformers.
\newblock In {\em CVPR}, 2021.

\bibitem{lin2021-mesh-graphormer}
Kevin Lin, Lijuan Wang, and Zicheng Liu.
\newblock Mesh graphormer.
\newblock In {\em ICCV}, 2021.

\bibitem{lin2014microsoft}
Tsung-Yi Lin, Michael Maire, Serge Belongie, James Hays, Pietro Perona, Deva
  Ramanan, Piotr Doll{\'a}r, and C~Lawrence Zitnick.
\newblock Microsoft coco: Common objects in context.
\newblock In {\em ECCV}, 2014.

\bibitem{liu2022temporal}
Zhenguang Liu, Runyang Feng, Haoming Chen, Shuang Wu, Yixing Gao, Yunjun Gao,
  and Xiang Wang.
\newblock Temporal feature alignment and mutual information maximization for
  video-based human pose estimation.
\newblock In {\em CVPR}, 2022.

\bibitem{liu2018large}
Ziwei Liu, Ping Luo, Xiaogang Wang, and Xiaoou Tang.
\newblock Deep learning face attributes in the wild.
\newblock 2015.

\bibitem{loper2015smpl}
Matthew Loper, Naureen Mahmood, Javier Romero, Gerard Pons-Moll, and Michael~J
  Black.
\newblock Smpl: A skinned multi-person linear model.
\newblock {\em ACM Trans. Graphics}, 2015.

\bibitem{Mahmood_2019_ICCV}
Naureen Mahmood, Nima Ghorbani, Nikolaus~F. Troje, Gerard Pons-Moll, and
  Michael~J. Black.
\newblock Amass: Archive of motion capture as surface shapes.
\newblock In {\em ICCV}, 2019.

\bibitem{jrdb}
Roberto Martin-Martin, Mihir Patel, Hamid Rezatofighi, Abhijeet Shenoi,
  JunYoung Gwak, Eric Frankel, Amir Sadeghian, and Silvio Savarese.
\newblock Jrdb: A dataset and benchmark of egocentric robot visual perception
  of humans in built environments.
\newblock {\em IEEE Trans. PAMI}, 2021.

\bibitem{mono-3dhp2017}
Dushyant Mehta, Helge Rhodin, Dan Casas, Pascal Fua, Oleksandr Sotnychenko,
  Weipeng Xu, and Christian Theobalt.
\newblock Monocular 3d human pose estimation in the wild using improved cnn
  supervision.
\newblock In {\em 3DV}, 2017.

\bibitem{moon2022accurate}
Gyeongsik Moon, Hongsuk Choi, and Kyoung~Mu Lee.
\newblock Accurate 3d hand pose estimation for whole-body 3d human mesh
  estimation.
\newblock In {\em CVPR}, 2022.

\bibitem{interhand}
Gyeongsik Moon, Shoou-I Yu, He Wen, Takaaki Shiratori, and Kyoung~Mu Lee.
\newblock Interhand2.6m: A dataset and baseline for 3d interacting hand pose
  estimation from a single rgb image.
\newblock In {\em ECCV}, 2020.

\bibitem{neverova2020continuous}
Natalia Neverova, David Novotny, Marc Szafraniec, Vasil Khalidov, Patrick
  Labatut, and Andrea Vedaldi.
\newblock Continuous surface embeddings.
\newblock In {\em NeurIPS}, 2020.

\bibitem{purushwalkam2020demystifying}
Senthil Purushwalkam and Abhinav Gupta.
\newblock Demystifying contrastive self-supervised learning: Invariances,
  augmentations and dataset biases.
\newblock In {\em NeurIPS}, 2020.

\bibitem{dpt}
Ren{\'e} Ranftl, Alexey Bochkovskiy, and Vladlen Koltun.
\newblock Vision transformers for dense prediction.
\newblock In {\em ICCV}, 2021.

\bibitem{rhodin2018unsupervised}
Helge Rhodin, Mathieu Salzmann, and Pascal Fua.
\newblock Unsupervised geometry-aware representation for 3d human pose
  estimation.
\newblock In {\em ECCV}, 2018.

\bibitem{rogez2015understanding}
Gr{\'e}gory Rogez, James~S Supancic, and Deva Ramanan.
\newblock Understanding everyday hands in action from rgb-d images.
\newblock In {\em ICCV}, 2015.

\bibitem{romero2022embodied}
Javier Romero, Dimitrios Tzionas, and Michael~J Black.
\newblock Embodied hands: Modeling and capturing hands and bodies together.
\newblock {\em ACM Trans. Graphics}, 2017.

\bibitem{Rothe-ICCVW-2015}
Rasmus Rothe, Radu Timofte, and Luc~Van Gool.
\newblock Dex: Deep expectation of apparent age from a single image.
\newblock In {\em ICCVW}, 2015.

\bibitem{imagenet}
Olga Russakovsky, Jia Deng, Hao Su, Jonathan Krause, Sanjeev Satheesh, Sean Ma,
  Zhiheng Huang, Andrej Karpathy, Aditya Khosla, Michael Bernstein,
  Alexander~C. Berg, and Li Fei-Fei.
\newblock {ImageNet Large Scale Visual Recognition Challenge}.
\newblock {\em IJCV}, 2015.

\bibitem{habitat}
Manolis Savva, Abhishek Kadian, Oleksandr Maksymets, Yili Zhao, Erik Wijmans,
  Bhavana Jain, Julian Straub, Jia Liu, Vladlen Koltun, Jitendra Malik, Devi
  Parikh, and Dhruv Batra.
\newblock Habitat: {A} {P}latform for {E}mbodied {AI} {R}esearch.
\newblock In {\em ICCV}, 2019.

\bibitem{schmidtke2021unsupervised}
Luca Schmidtke, Athanasios Vlontzos, Simon Ellershaw, Anna Lukens, Tomoki
  Arichi, and Bernhard Kainz.
\newblock Unsupervised human pose estimation through transforming shape
  templates.
\newblock In {\em CVPR}, 2021.

\bibitem{sun2021monocular}
Yu Sun, Qian Bao, Wu Liu, Yili Fu, Michael~J Black, and Tao Mei.
\newblock Monocular, one-stage, regression of multiple 3d people.
\newblock In {\em CVPR}, 2021.

\bibitem{tang2023humanbench}
Shixiang Tang, Cheng Chen, Qingsong Xie, Meilin Chen, Yizhou Wang, Yuanzheng
  Ci, Lei Bai, Feng Zhu, Haiyang Yang, Li Yi, et~al.
\newblock Humanbench: Towards general human-centric perception with projector
  assisted pretraining.
\newblock In {\em CVPR}, 2023.

\bibitem{videomae}
Zhan Tong, Yibing Song, Jue Wang, and Limin Wang.
\newblock Videomae: Masked autoencoders are data-efficient learners for
  self-supervised video pre-training.
\newblock In {\em NeurIPS}, 2022.

\bibitem{aist}
Shuhei Tsuchida, Satoru Fukayama, Masahiro Hamasaki, and Masataka Goto.
\newblock Aist dance video database: Multi-genre, multi-dancer, and
  multi-camera database for dance information processing.
\newblock In {\em ISMIR}, 2019.

\bibitem{van2018inaturalist}
Grant Van~Horn, Oisin Mac~Aodha, Yang Song, Yin Cui, Chen Sun, Alex Shepard,
  Hartwig Adam, Pietro Perona, and Serge Belongie.
\newblock The inaturalist species classification and detection dataset.
\newblock In {\em CVPR}, 2018.

\bibitem{Varol_2017_CVPR}
Gul Varol, Javier Romero, Xavier Martin, Naureen Mahmood, Michael~J. Black,
  Ivan Laptev, and Cordelia Schmid.
\newblock Learning from synthetic humans.
\newblock In {\em CVPR}, 2017.

\bibitem{3dpw}
Timo von Marcard, Roberto Henschel, Michael~J. Black, Bodo Rosenhahn, and
  Gerard Pons-Moll.
\newblock Recovering accurate 3d human pose in the wild using imus and a moving
  camera.
\newblock In {\em ECCV}, 2018.

\bibitem{maskfeat}
Chen Wei, Haoqi Fan, Saining Xie, Chao-Yuan Wu, Alan Yuille, and Christoph
  Feichtenhofer.
\newblock {Masked Feature Prediction for Self-Supervised Visual Pre-Training}.
\newblock In {\em CVPR}, 2022.

\bibitem{croco}
{Weinzaepfel, Philippe and Leroy, Vincent and Lucas, Thomas and Br\'egier,
  Romain and Cabon, Yohann and Arora, Vaibhav and Antsfeld, Leonid and
  Chidlovskii, Boris and Csurka, Gabriela and Revaud J\'er\^ome}.
\newblock {CroCo: Self-Supervised Pre-training for 3D Vision Tasks by
  Cross-View Completion}.
\newblock In {\em {NeurIPS}}, 2022.

\bibitem{wu2018npid}
Zhirong Wu, Yuanjun Xiong, Stella~X. Yu, and Dahua Lin.
\newblock Unsupervised feature learning via non-parametric instance-level
  discrimination.
\newblock In {\em CVPR}, 2018.

\bibitem{SimMIM}
Zhenda Xie, Zheng Zhang, Yue Cao, Yutong Lin, Jianmin Bao, Zhuliang Yao, Qi
  Dai, and Han Hu.
\newblock {SimMIM: A Simple Framework for Masked Image Modeling}.
\newblock In {\em CVPR}, 2022.

\bibitem{texformer}
Xiangyu Xu and Chen~Change Loy.
\newblock {3D} human texture estimation from a single image with transformers.
\newblock In {\em ICCV}, 2021.

\bibitem{vitpose}
Yufei Xu, Jing Zhang, Qiming Zhang, and Dacheng Tao.
\newblock Vitpose: Simple vision transformer baselines for human pose
  estimation.
\newblock In {\em NeurIPS}, 2022.

\bibitem{yu2015lsun}
Fisher Yu, Ari Seff, Yinda Zhang, Shuran Song, Thomas Funkhouser, and Jianxiong
  Xiao.
\newblock Lsun: Construction of a large-scale image dataset using deep learning
  with humans in the loop.
\newblock {\em arXiv preprint arXiv:1506.03365}, 2015.

\bibitem{humbi}
Zhixuan Yu, Jae~Shin Yoon, In~Kyu Lee, Prashanth Venkatesh, Jaesik Park, Jihun
  Yu, and Hyun~Soo Park.
\newblock Humbi: A large multiview dataset of human body expressions.
\newblock In {\em CVPR}, 2020.

\bibitem{lpips}
R. Zhang, P. Isola, A.~A. Efros, E. Shechtman, and O. Wang.
\newblock The unreasonable effectiveness of deep features as a perceptual
  metric.
\newblock In {\em CVPR}, 2018.

\bibitem{pennaction}
Weiyu Zhang, Menglong Zhu, and Konstantinos~G. Derpanis.
\newblock From actemes to action: A strongly-supervised representation for
  detailed action understanding.
\newblock In {\em ICCV}, 2013.

\bibitem{zhao2021learning}
Long Zhao, Yuxiao Wang, Jiaping Zhao, Liangzhe Yuan, Jennifer~J Sun, Florian
  Schroff, Hartwig Adam, Xi Peng, Dimitris Metaxas, and Ting Liu.
\newblock Learning view-disentangled human pose representation by contrastive
  cross-view mutual information maximization.
\newblock In {\em CVPR}, 2021.

\bibitem{mars}
Liang Zheng, Zhi Bie, Yifan Sun, Jingdong Wang, Chi Su, Shengjin Wang, and Qi
  Tian.
\newblock Mars: A video benchmark for large-scale person re-identification.
\newblock In {\em ECCV}, 2016.

\bibitem{hanco}
Christian Zimmermann, Max Argus, and Thomas Brox.
\newblock Contrastive representation learning for hand shape estimation.
\newblock In {\em GCPR}, 2021.

\bibitem{zimmermann2019freihand}
Christian Zimmermann, Duygu Ceylan, Jimei Yang, Bryan Russell, Max Argus, and
  Thomas Brox.
\newblock Freihand: A dataset for markerless capture of hand pose and shape
  from single rgb images.
\newblock In {\em ICCV}, 2019.

\end{thebibliography}
}

\newpage
\appendix

{\huge \bf{Appendix} \vspace{0.5cm}}

This appendix presents additional results on the human texture estimation task (Section~\ref{sec:texture}), training cost information (Section~\ref{sec:timing}) and qualitative results of the pre-training objective as well as of several downstream tasks (Section~\ref{sec:vizu}).

\section{Human texture estimation}
\label{sec:texture}
Our pre-training objective has some similarities with the task of novel-view synthesis. Given an observation of a person (the reference image), and some information about a target pose and viewpoint (the masked target image), the network is trained to reconstruct an image of the person from said viewpoint.
In order to evaluate this particular facet of human understanding, we compare different pre-training strategies on the task of human novel-view generation.
\begin{figure}[!h]
    \centering
    \includegraphics[width=\linewidth]{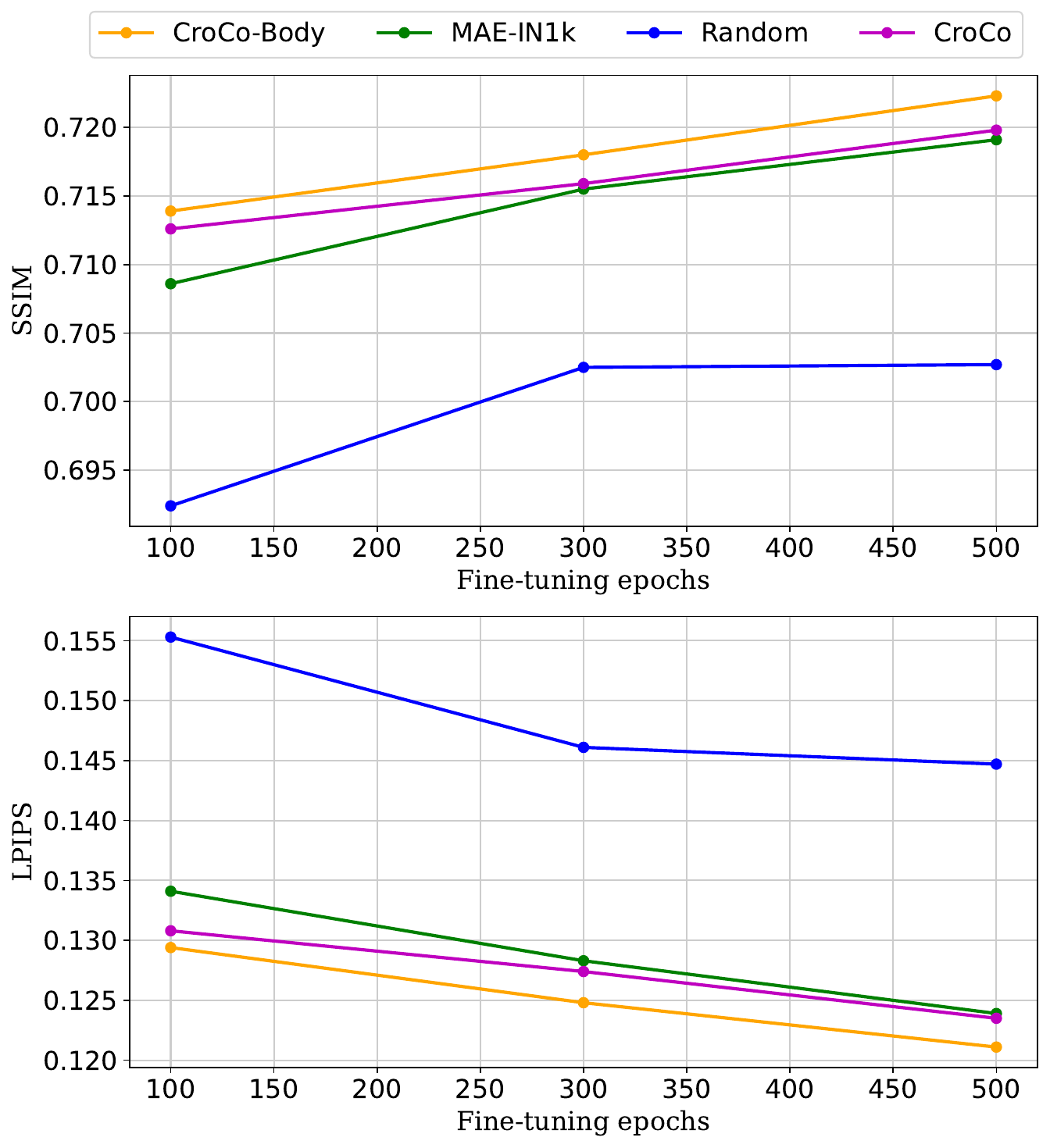} \\[-0.3cm]
    \caption{\textbf{Evaluation scores of various pre-trained models on the texture estimation task of TexFormer~\cite{texformer}}, at different fine-tuning stages.
    From left to right, we report SSIM$\uparrow$ (structural similarity index) and LPIPS$\downarrow$~\cite{lpips} metrics. All models return a single RGB texture.}
    \label{fig:texformer}
\end{figure}
More particularly, we tackle human texture generation from a single image, following the experimental setup of TexFomer~\cite{texformer}. They define a key, query and value images which are partly pre-computed, and partly based on the input image. These images are encoded at different scales using $3$ CNNs, then transformer layers perform multi-headed attention at different scales. Resulting features are merged through another CNN.
We modify their code, replacing their whole network with our ViT-based encoder-decoder architecture. The value image is discarded, and encoder weights are fine-tuned independently for key and query images. The network is trained to return a single RGB texture.
This adaptation is a bit naive, but our goal is mainly to compare different pre-training methods on a different task, that leverages both encoder and decoder of the pre-trained network.
We follow the TexFormer experimental setup in terms of hyperparameters, datasets, and metrics. 
Results for different network initializations are reported in Figure~\ref{fig:texformer}. For MAE, we randomly initialize the decoder weights.
Pre-training the model does help a lot: both CroCo and MAE provide a significant boost. CroCo performs slightly better, which is probably due in part to the pre-trained decoder.
\OursBody outperforms both CroCo and MAE.

\begin{figure*}[!ht]
\centering
\newcommand{\prewidth}{0.14\linewidth}
\begin{tabular}{P{\prewidth}P{\prewidth}P{\prewidth}P{\prewidth}P{\prewidth}P{\prewidth}}
Reference & Masked & \multirow{2}{*}{CroCo} & \OursBody  & \multirow{2}{*}{\OursBody} & Target \\
view        &      input           &       & (no ref)  &           &  image
\end{tabular} \vspace{-0.3cm}
\includegraphics[width=\linewidth]{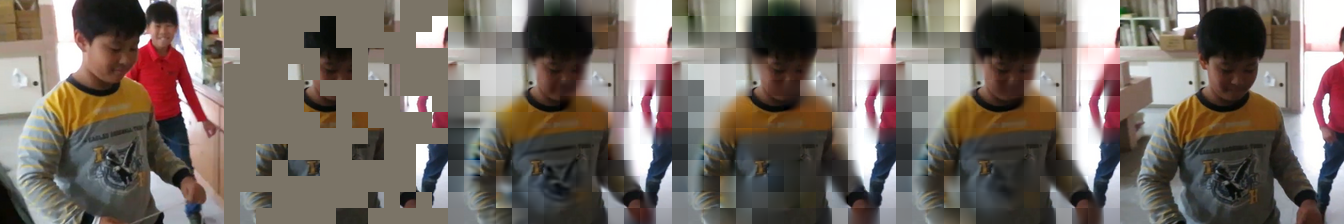} \\
\includegraphics[width=\linewidth]{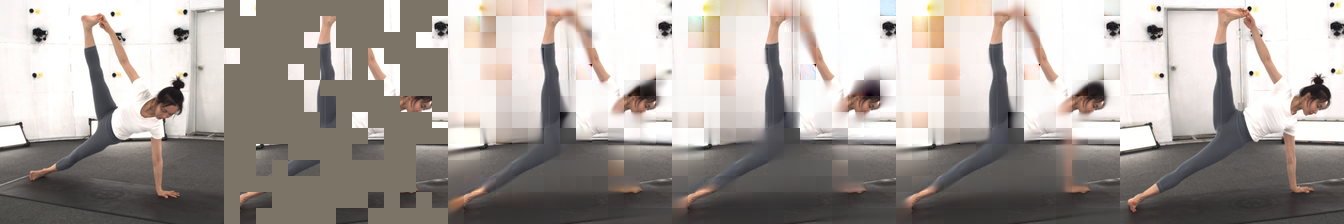} \\
\includegraphics[width=\linewidth]{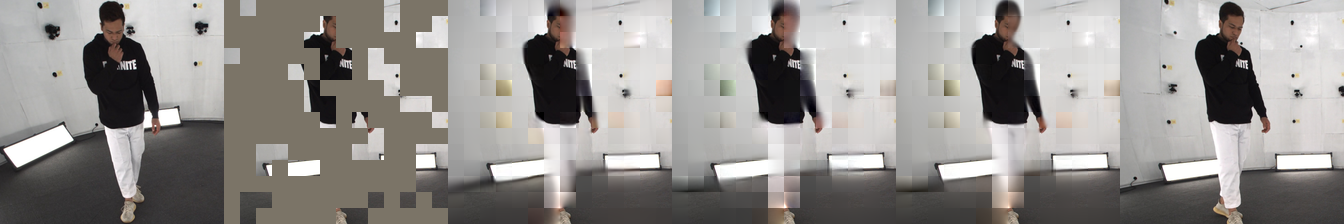} \\[-0.2cm]
\caption{\textbf{Completion examples on cross-view (\ie multi-view) pairs} from the Mannequin Challenge dataset~\cite{mannequinchallenge} (first row) and the GeneBody dataset~\cite{genebody} (last two rows).
\OursBody (no ref) stands for our model evaluated on the masked input, and a reference view set to zero (\ie, a fully-black image).
}
\label{fig:pretrain_multiview}
\end{figure*}
\begin{figure*}[!h]
\centering
\newcommand{\prewidth}{0.14\linewidth}
\begin{tabular}{P{\prewidth}P{\prewidth}P{\prewidth}P{\prewidth}P{\prewidth}P{\prewidth}}
Reference & Masked & \multirow{2}{*}{CroCo} & \OursBody  & \multirow{2}{*}{\OursBody} & Target \\
view        &      input           &       & (no ref)  &           &  image
\end{tabular} \vspace{-0.3cm}
\includegraphics[width=\linewidth]{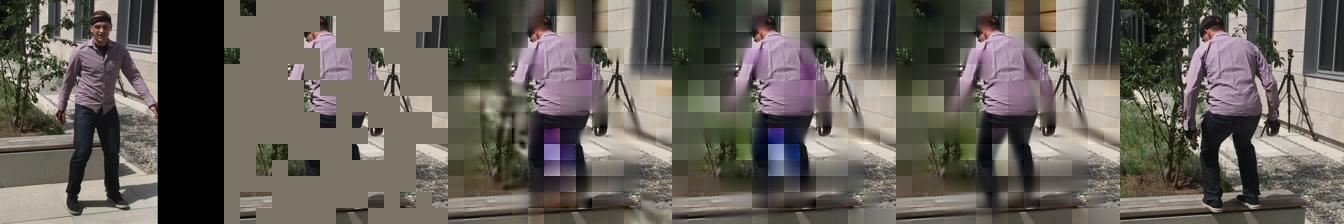} \\
\includegraphics[width=\linewidth]{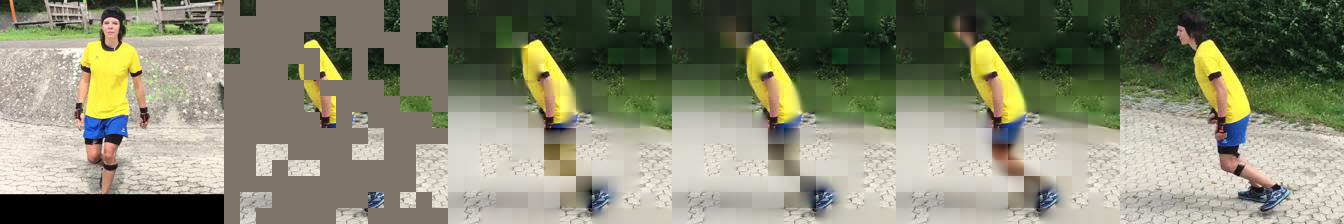} \\
\caption{\textbf{Completion examples on cross-pose (\ie temporal) pairs} from 3DPW~\cite{3dpw} validation set (unseen during pre-training).
\OursBody (no ref) stands for our model evaluated on the masked input, and a reference view set to zero (\ie, a fully-black image).
}
\label{fig:pretrain_temporal}
\end{figure*}

\section{Training time}
\label{sec:timing}

In this section, we give timings necessary for pre-training and fine-tuning our models.
Pre-training the CroCo-Body model takes about 8 days on 4 NVIDIA A100 GPUs.
Fine-tuning it on a single A100 takes about half a day per downstream task.
As for the CroCo-Hand model, pre-training on 4 V100 GPUs requires 2.25 days, and fine-tuning on a single V100 GPU takes about 8 hours per downstream task.

\section{Qualitative results}
\label{sec:vizu}
\subsection{Pre-training}

\noindent \textbf{\OursBody.}
We illustrate the pre-training task of \OursBody on both cross-view and cross-pose pairs in Figures~\ref{fig:pretrain_multiview} and~\ref{fig:pretrain_temporal} respectively, 
with data never seen by the model during pre-training.
We report predictions of \OursBody using either the reference image or a reference image entirely black (`no ref'), to ablate the cross-image completion capabilities of the decoder.
CroCo tends to recover
detailed patterns on relatively flat surfaces, such as the t-shirt logo on the first row of Figure~\ref{fig:pretrain_multiview}. It lacks prior knowledge about humans however, and struggles to reconstruct the left arm on the second row. In contrast, \OursBody produces a sharper arm reconstruction, which may be attributed to its human-specific pre-training and the ability to leverage the reference view. A similar effect is visible on the reconstruction of the head in the last row.

For cross-pose pairs (Figure~\ref{fig:pretrain_temporal}), we observe that completions of CroCo are similar to the ones of \OursBody with no reference image. This suggests that CroCo benefits little from cross-image attention, being specifically trained to exploit static stereoscopic pairs only.
\OursBody on the other hand seems able to recover information from the reference image about the lower-body garments even though they are heavily occluded in the masked target in both examples, and achieves a better completion of the masked image.

\begin{figure}[t]
\captionsetup[subfigure]{labelformat=empty}
\centering
\newcommand{\prewidth}{0.14\linewidth}
\begin{subfigure}{0.24\linewidth}
\caption{Reference \\ view}
\includegraphics[width=\linewidth]{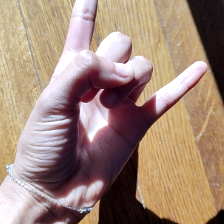}
\end{subfigure}
\begin{subfigure}{0.24\linewidth}
\caption{Masked \\ input}
\includegraphics[width=\linewidth]{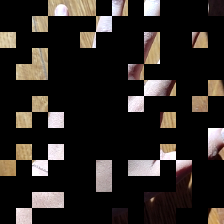}
\end{subfigure}
\begin{subfigure}{0.24\linewidth}
\caption{\OursHand}
\includegraphics[width=\linewidth]{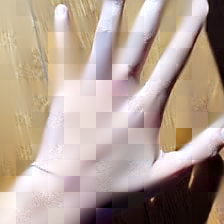}
\end{subfigure}
\begin{subfigure}{0.24\linewidth}
\caption{Target \\ image}
\includegraphics[width=\linewidth]{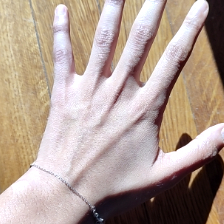}
\end{subfigure}
\begin{subfigure}{0.24\linewidth}
\includegraphics[width=\linewidth]{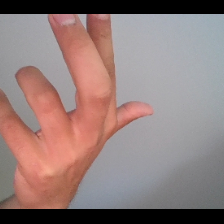}
\end{subfigure}
\begin{subfigure}{0.24\linewidth}
\includegraphics[width=\linewidth]{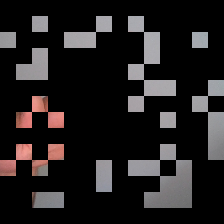}
\end{subfigure}
\begin{subfigure}{0.24\linewidth}
\includegraphics[width=\linewidth]{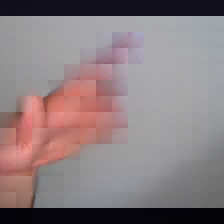}
\end{subfigure}
\begin{subfigure}{0.24\linewidth}
\includegraphics[width=\linewidth]{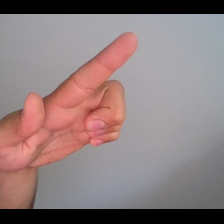}
\end{subfigure}
\begin{subfigure}{0.24\linewidth}
\includegraphics[width=\linewidth]{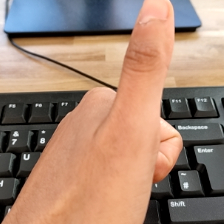}
\end{subfigure}
\begin{subfigure}{0.24\linewidth}
\includegraphics[width=\linewidth]{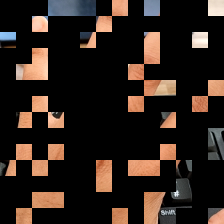}
\end{subfigure}
\begin{subfigure}{0.24\linewidth}
\includegraphics[width=\linewidth]{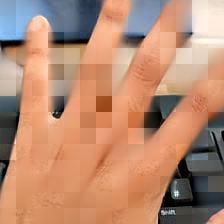}
\end{subfigure}
\begin{subfigure}{0.24\linewidth}
\includegraphics[width=\linewidth]{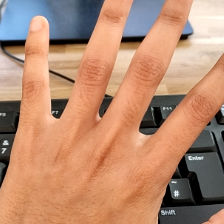}
\end{subfigure}
\vspace{-0.3cm}
\caption{\textbf{Completion examples of \OursHand on unseen cross-pose (\ie temporal) pairs}, in indoor scenes.}
\vspace{-0.2cm}
\label{fig:pretrain_crocohand}
\end{figure}
\begin{figure}[t]
\captionsetup[subfigure]{labelformat=empty}
\centering
\newcommand{\prewidth}{0.14\linewidth}
\begin{subfigure}{0.24\linewidth}
\caption{Reference \\ view}
\includegraphics[width=\linewidth]{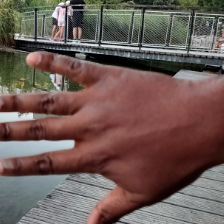}
\end{subfigure}
\begin{subfigure}{0.24\linewidth}
\caption{Masked \\ input}
\includegraphics[width=\linewidth]{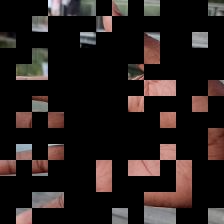}
\end{subfigure}
\begin{subfigure}{0.24\linewidth}
\caption{\OursHand}
\includegraphics[width=\linewidth]{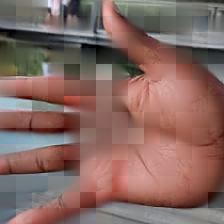}
\end{subfigure}
\begin{subfigure}{0.24\linewidth}
\caption{Target \\ image}
\includegraphics[width=\linewidth]{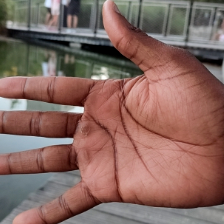}
\end{subfigure}
\begin{subfigure}{0.24\linewidth}
\includegraphics[width=\linewidth]{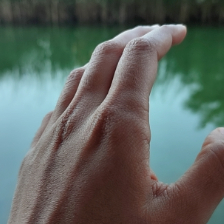}
\end{subfigure}
\begin{subfigure}{0.24\linewidth}
\includegraphics[width=\linewidth]{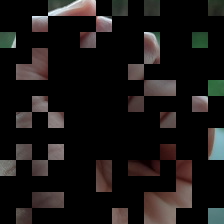}
\end{subfigure}
\begin{subfigure}{0.24\linewidth}
\includegraphics[width=\linewidth]{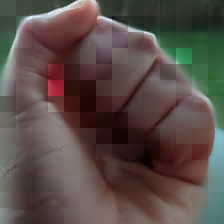}
\end{subfigure}
\begin{subfigure}{0.24\linewidth}
\includegraphics[width=\linewidth]{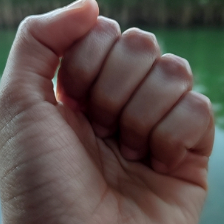}
\end{subfigure}
\begin{subfigure}{0.24\linewidth}
\includegraphics[width=\linewidth]{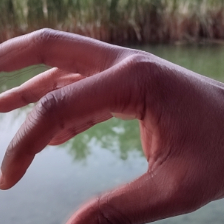}
\end{subfigure}
\begin{subfigure}{0.24\linewidth}
\includegraphics[width=\linewidth]{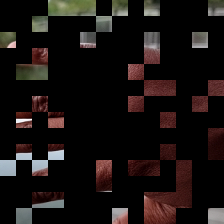}
\end{subfigure}
\begin{subfigure}{0.24\linewidth}
\includegraphics[width=\linewidth]{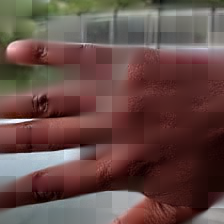}
\end{subfigure}
\begin{subfigure}{0.24\linewidth}
\includegraphics[width=\linewidth]{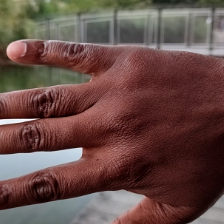}
\end{subfigure}
\vspace{-0.3cm}
\caption{\textbf{Completion examples of \OursHand on unseen cross-pose (\ie temporal) pairs}, in outdoor scenes.}
\vspace{-0.2cm}
\label{fig:pretrain_crocohand1}
\end{figure}

\noindent \textbf{\OursHand.}
We illustrate the pre-training task of \OursHand on unseen cross-pose pairs in indoor and outdoor scenes in Figures~\ref{fig:pretrain_crocohand} and~\ref{fig:pretrain_crocohand1}, respectively.
We tested \OursHand on internal images which have never been seen during the pre-training stage.
We observe that \OursHand learned the structure of a human hand such as shown in Figure~\ref{fig:pretrain_crocohand} where it reconstructs a pointed index finger from a small handful of visible palm patches.
\OursHand also performs well on outdoor images such as shown in Figure~\ref{fig:pretrain_crocohand1}, despite the fact the pre-training is done integrally using data captured in labs.
It is also interesting to notice that \OursHand also generalizes well to different skin tones.

\begin{figure}[h]
\captionsetup[subfigure]{labelformat=empty}
    \centering
    \begin{subfigure}{0.24\linewidth} \caption{\centering Input \protect\linebreak  image} \includegraphics[width=\linewidth]{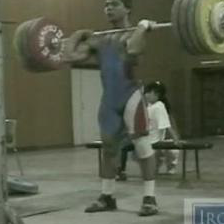}  \end{subfigure}
    \begin{subfigure}{0.24\linewidth} \caption{\centering Predicted \protect\linebreak  heatmap} \includegraphics[width=\linewidth]{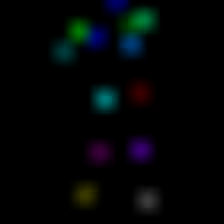} \end{subfigure}
    \begin{subfigure}{0.24\linewidth} \caption{\centering Ground-truth \protect\linebreak  heatmap} \includegraphics[width=\linewidth]{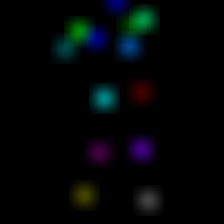}  \end{subfigure}
    \begin{subfigure}{0.24\linewidth} \caption{\centering Predicted \protect\linebreak   keypoints} \includegraphics[width=\linewidth]{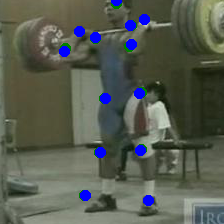}  \end{subfigure}
    \vspace{-0.2cm}
    \caption{\textbf{Visualization of the supervised pretext task used for the ablation in Table 3 of the main paper}. The right shows the position of predicted keypoints (blue) obtained with a simple argmax on the predicted heatmap, on top of ground truth keypoints (green). Heatmaps have been artificially converted to 3-channels images for visualization purpose.}
    \label{fig:kpts_viz}
\end{figure}

\noindent \textbf{Keypoints supervision.}
We give here more detailed information about the keypoints supervision used for the pre-training ablation in Section 4.2 and Table 3 of the main paper.
We select the set of 13 keypoints used in PennAction~\cite{pennaction}. For each pre-training image, we generate a 13-channels keypoint heatmap where each keypoint is represented as a Gaussian with $\sigma=8$ pixels.
Figure~\ref{fig:kpts_viz} illustrates the task on a simple example.
During pre-training, the encoded image is passed through a simple prediction head that is trained to predict the heatmaps with a simple binary cross-entropy loss. Ground-truth keypoints are weighted according to a confidence parameter ($0$ for missing keypoints). When pre-training with both objectives (Table 3 of the main paper, last row), we train the keypoints prediction on the encoded reference image, that is fully visible.

\subsection{Downstream results}
\begin{figure}[h]
\centering
\newcommand{\DPreswidth}{0.24\linewidth}
\newcommand{\RGBreswidth}{0.24\linewidth}
\captionsetup[subfigure]{labelformat=empty}

\begin{subfigure}{\DPreswidth} \caption{\centering Input \protect\linebreak  image} \includegraphics[width=\textwidth]{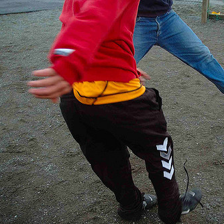} \end{subfigure}
\begin{subfigure}{\DPreswidth} \caption{\centering Predicted \protect\linebreak  UV coordinates} \includegraphics[width=\textwidth]{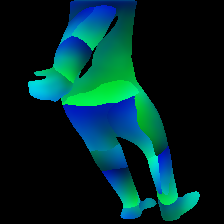}\end{subfigure}
\begin{subfigure}{\DPreswidth} \caption{\centering Predicted \protect\linebreak  labels} \includegraphics[width=\textwidth]{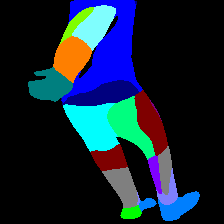} \end{subfigure}
\begin{subfigure}{\DPreswidth} \caption{\centering Ground-truth \protect\linebreak  labels} \includegraphics[width=\textwidth]{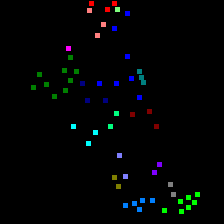} \end{subfigure} \\

\begin{subfigure}{\DPreswidth} \includegraphics[width=\textwidth]{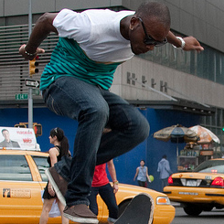}  \end{subfigure}
\begin{subfigure}{\DPreswidth} \includegraphics[width=\textwidth]{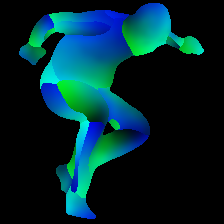}  \end{subfigure}
\begin{subfigure}{\DPreswidth} \includegraphics[width=\textwidth]{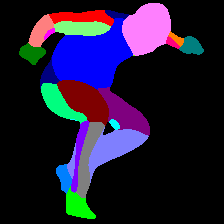}  \end{subfigure}
\begin{subfigure}{\DPreswidth} \includegraphics[width=\textwidth]{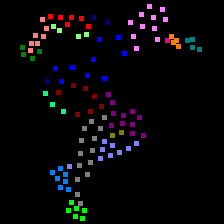}  \end{subfigure}
\vspace{-0.3cm}
\caption{\textbf{Qualitative results of \OursBody on the DensePose task} on the COCO dataset. The sparse ground-truth labels used for training and evaluation are dilated here for visualization purposes.
}
\label{fig:quali_dp}
\end{figure}

\begin{figure}[h]
\centering
\begin{subfigure}[b]{1\linewidth}
\includegraphics[width=0.49\linewidth]{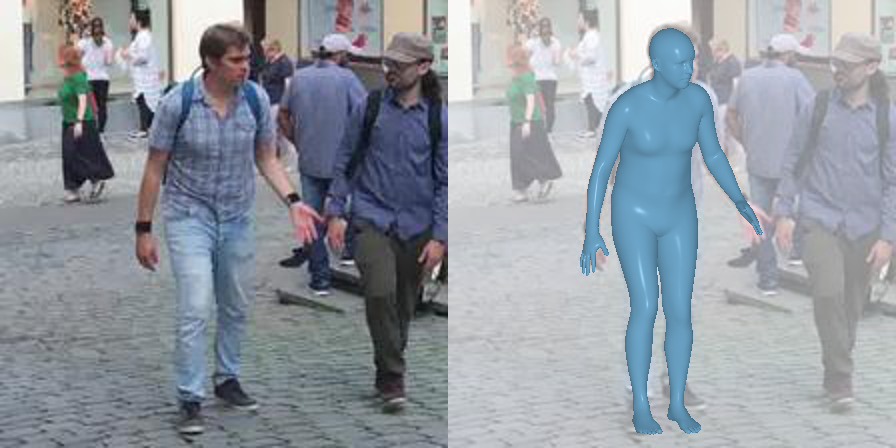} 
\includegraphics[width=0.49\linewidth]{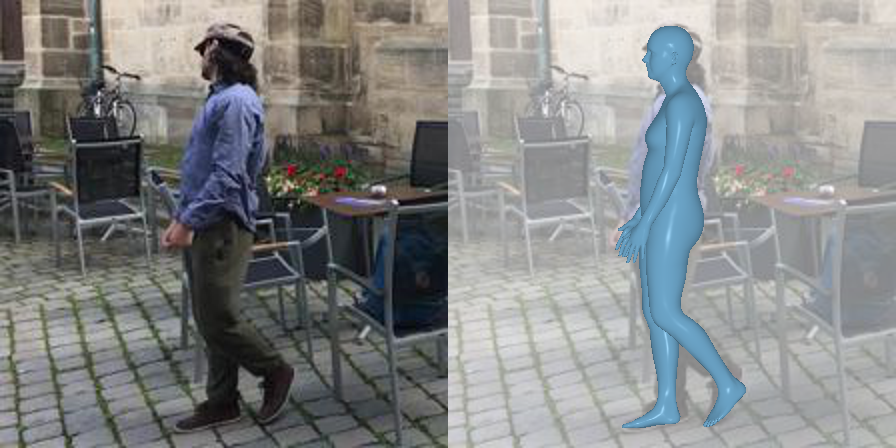}
\caption{\textbf{Results on the body mesh recovery task} on 3DPW~\cite{3dpw}.}
\end{subfigure}
\\
\begin{subfigure}[b]{1\linewidth}
\includegraphics[width=0.49\linewidth]{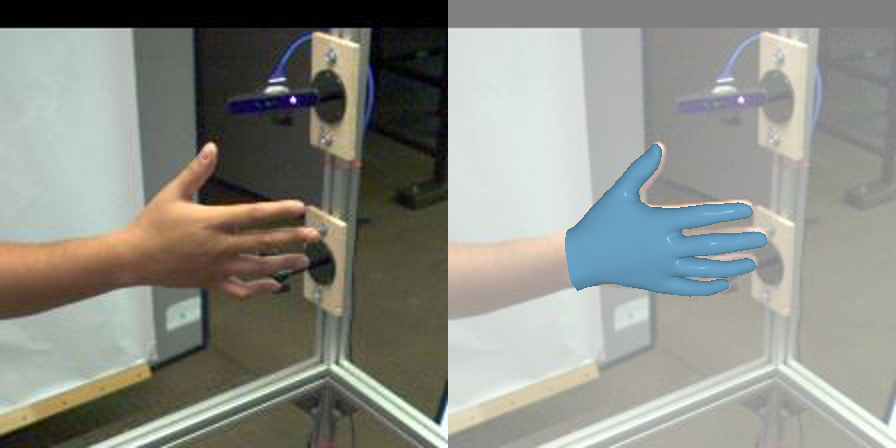} 
\includegraphics[width=0.49\linewidth]{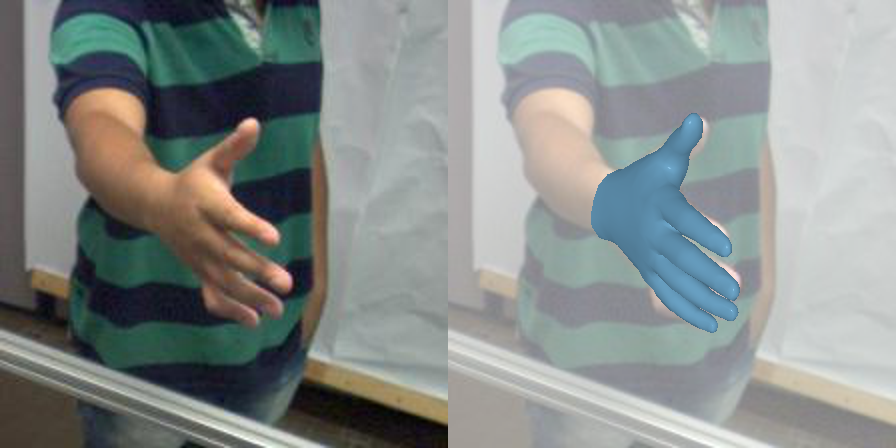}
\caption{\textbf{Results on the hand mesh recovery task} on HanCo~\cite{hanco}.}
\end{subfigure}
\vspace{-0.65cm}
\caption{\textbf{Qualitative examples of our models on the two mesh recovery tasks.} Each pair shows the input image, and the output of \OursBody (top) or \OursHand (bottom), overlaid on the image.}
\label{fig:quali_bmr}
\end{figure}

\begin{figure}[h]
\centering
\resizebox{\linewidth}{!}{
\newcommand{\figclassifwidth}{0.24\textwidth}
\begin{tabular}{c@{ }c@{ }c@{ }c@{ }c@{ }c}
\includegraphics[width=\figclassifwidth]{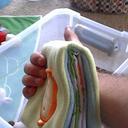} &
\includegraphics[width=\figclassifwidth]{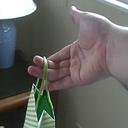} &
\includegraphics[width=\figclassifwidth]{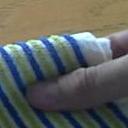} &
\includegraphics[width=\figclassifwidth]{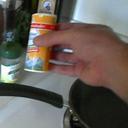}
\\[-0.1cm]
\textcolor{ForestGreen}{GT: Palmar} & 
\textcolor{ForestGreen}{GT: Fixed Hook} &
\textcolor{ForestGreen}{GT: Parallel Extension} &
\textcolor{ForestGreen}{GT: Prismatic 4 Finger}
\\[0.07cm]
Palmar: \textcolor{ForestGreen}{96.46\%} &
Fixed Hook: \textcolor{ForestGreen}{99.72\%} &
Palmar: 99.99\% &
Prismatic 4 Finger: \textcolor{ForestGreen}{96.41\%}
\\
Adducted Thumb: 1.83\% &
Distal Type: 0.24\% &
Medium Wrap: 0.01\% &
Precision Sphere: 1.44\%
\\
Medium Wrap: 0.51\% &
Precision Disk: 0.02\% &
Adducted Thumb: 0.00\% &
Sphere 4 Finger: 1.27\%
\end{tabular}
}
\vspace{-0.2cm}
\caption{\textbf{Qualitative examples of our models on the grasp classification task} on GUN-71~\cite{rogez2015understanding}. For the images on the top row, we show below the ground-truth class as well as the top 3 prediction made by our model.}
\label{fig:classif_grasp}
\end{figure}

We now show some visualizations of the different downstream tasks that we evaluate on. Figures~\ref{fig:quali_dp} and~\ref{fig:quali_bmr} show results on regression tasks (DensePose and body/hand mesh recovery, respectively), while Figure~\ref{fig:classif_grasp} shows results on the grasp classification task.

\end{document}